\pdfoutput=1
\documentclass[11pt,a4paper]{article}
\usepackage[bottom]{footmisc}
\usepackage{emnlp2021}
\usepackage{times}
\usepackage{latexsym}
\usepackage[T1]{fontenc}
\usepackage[utf8]{inputenc}

\usepackage{microtype}

\usepackage{graphicx}	
\usepackage{balance}
\usepackage{amsfonts,amsmath,amssymb,amsthm}
\usepackage{bm}

\usepackage{booktabs} 
\usepackage{threeparttable}
\usepackage[htt]{hyphenat}
\usepackage{enumitem}
\usepackage{xspace}
\usepackage{multirow}
\usepackage{hhline}
\usepackage{hyperref}
\usepackage{url}
\usepackage{colortbl}
\usepackage{subfig}
\usepackage[normalem]{ulem} 
\usepackage{algorithm}
\usepackage{appendix}
\usepackage[noend]{algorithmic}

\newcommand{\ignore}[1]{}

\newcommand{\ie}{\textit{i.e.}\xspace}
\newcommand{\viz}{\textit{viz.}\xspace}
\newcommand{\eg}{\textit{e.g.}\xspace}
\newcommand{\cf}{\textit{cf.}\xspace}

\newcommand{\vs}{\textit{vs.}\xspace}
\newcommand{\etc}{\textit{etc.}\xspace}

\newcommand{\ethemes}{\textsc{Eigenthemes}\xspace}
\newcommand{\wethemes}{\textsc{Weighted Eigenthemes}\xspace}

\newcommand{\dw}{\textsc{Deepwalk}\xspace}
\newcommand{\wv}{\textsc{Word2vec}\xspace}
\newcommand{\nm}{\textsc{NameMatch}\xspace}
\newcommand{\degree}{\textsc{Degree}\xspace}
\newcommand{\lctxt}{\textsc{Local Ctxt}\xspace}
\newcommand{\gctxt}{\textsc{Global Ctxt}\xspace}
\newcommand{\avg}{\textsc{Avg}\xspace}

\newcommand{\eshort}{\textsc{Eigen}\xspace}

\newcommand{\titov}{$\tau$MIL-ND\xspace}
\newcommand{\wtitov}{W$\tau$MIL-ND\xspace}

\newcommand{\xhdr}[1]{\vspace{1.3mm}\noindent{{\bf #1.}}}
\newcommand{\xhdrNoPeriod}[1]{\vspace{1.7mm}\noindent{{\bf #1}}}

\newcommand{\moveup}{\vspace*{-2mm}}
\newcommand{\moveups}{\vspace*{-1mm}}
\newcommand{\tabcaption}[1]{\vspace*{-3mm}\caption{#1}\vspace*{-4mm}}
\newcommand{\figcaption}[1]{\vspace*{-3mm}\caption{#1}\vspace*{-5mm}}

\title{Low-Rank Subspaces for Unsupervised Entity Linking}

\author{
  Akhil Arora \\
  EPFL \\
  \texttt{akhil.arora@epfl.ch} \\\And
  Alberto Garc\'{i}a-Dur\'{a}n\thanks{~~Research done while at EPFL.} \\
  Atinary Technologies\\
  \texttt{agaduran@gmail.com} \\\And
  Robert West \\
  EPFL\\
  \texttt{robert.west@epfl.ch} \\
}

\begin{document}
\maketitle

\begin{abstract}
Entity linking is an important problem with many applications.
Most previous solutions were designed for settings where annotated training data is available, which is, however, not the case in numerous domains.
We propose a light-weight and scalable entity linking method, \ethemes, that relies solely on the availability of entity names and a referent knowledge base.
\ethemes exploits the fact that the entities that are truly mentioned in a document (the ``gold entities'') tend to form a semantically dense subset of the set of all candidate entities in the document.
Geometrically speaking, when representing entities as vectors via some given embedding, the gold entities tend to lie in a low-rank subspace of the full embedding space.
\ethemes identifies this subspace using the singular value decomposition and scores candidate entities according to their proximity to the subspace. 
On the empirical front, we introduce multiple strong baselines that compare favorably to (and sometimes even outperform) the existing state of the art. 
Extensive experiments on benchmark datasets from a variety of real-world domains showcase the effectiveness of our approach.
\end{abstract}

\section{Introduction}
\label{sec:intro}
Entity linking (EL) is the task of grounding mentions to a reference knowledge base (also referred to as knowledge graph). With a plethora of applications, including but not limited to information extraction \cite{hoffart} and automatic knowledge base construction \cite{gao2018building}, EL is one of the most actively researched topics in natural language processing. 
Despite the recent proliferation of EL methods, recent works \cite{xel3,xel4} have pointed out that the performance of existing techniques largely relies on the existence of large corpora of annotated data. 

\begin{figure*}
    \centering
    \moveup
    \moveup
    \includegraphics[width=0.62\textwidth]{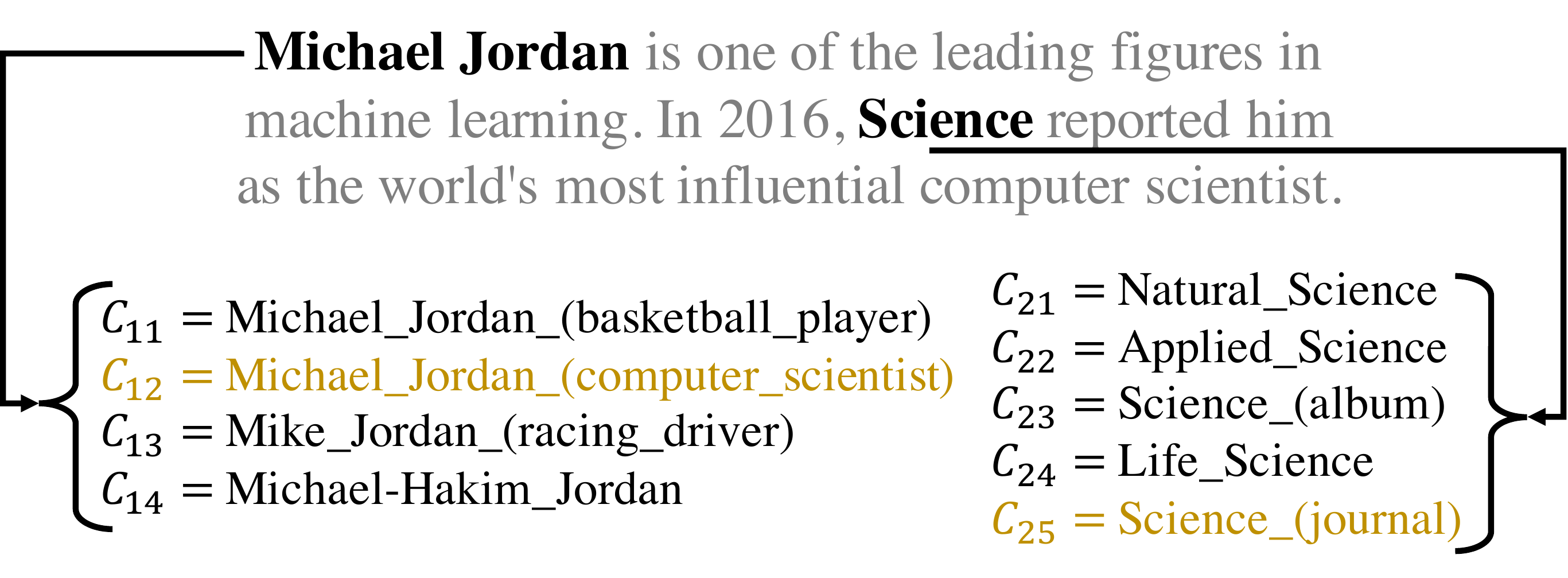}
    \hfill
    \includegraphics[trim=2.2cm 0.9cm 0.7cm 0cm, width=0.27\textwidth]{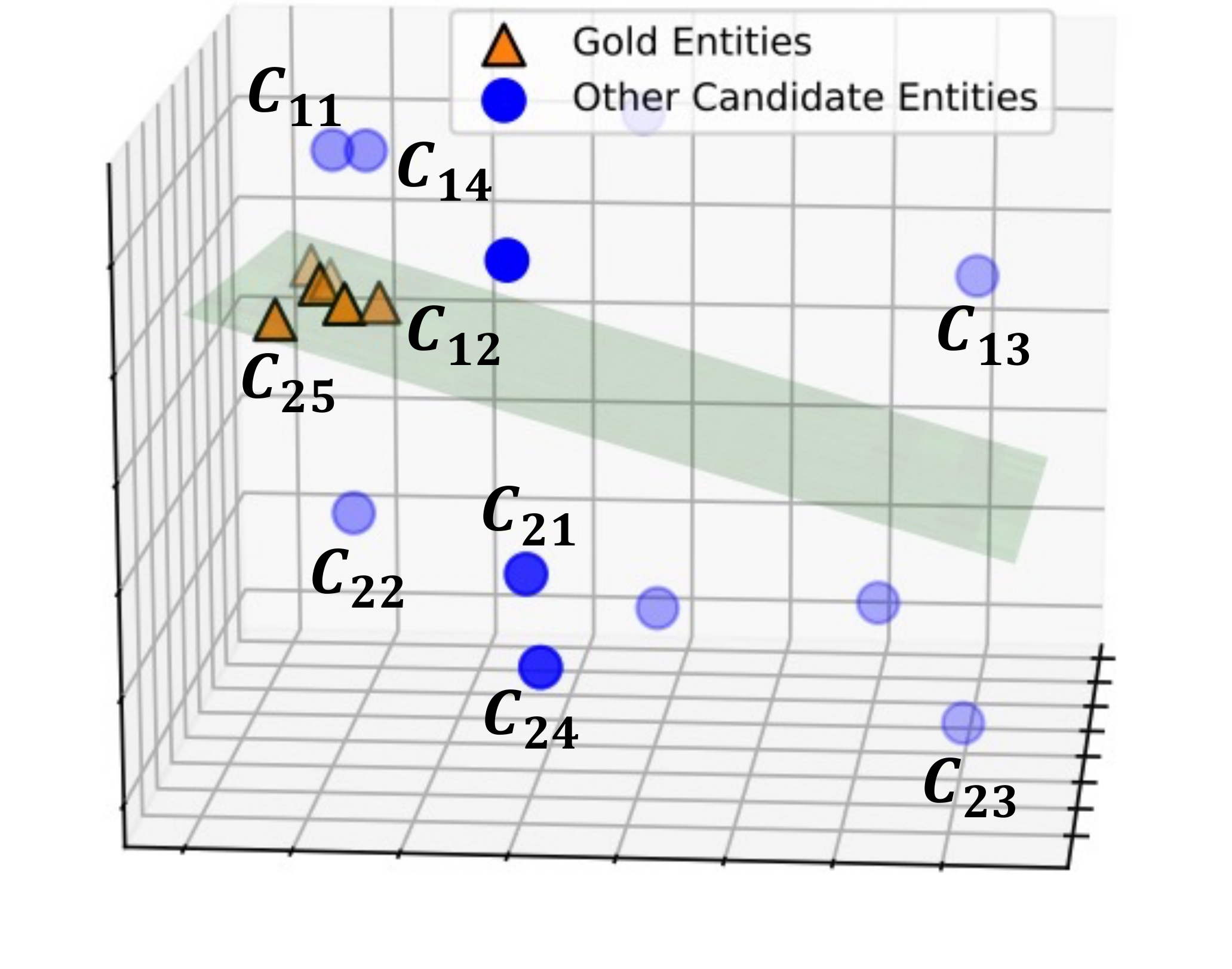}
    \vspace*{1mm}
    \figcaption{(Left) Excerpt from a Wikipedia article about \emph{Michael I. Jordan} with two mentions (\textbf{in bold}) and corresponding candidate entities generated using surface name matching. (Right) Embedding space for all the candidate entities within a document. The gold entities (triangles) form a cluster in the proximity of the subspace (green plane) identified by \ethemes, while other candidates (circles) are distant from the subspace.}
    \label{fig:motivation}
\end{figure*}

\xhdr{Unsupervised entity linking}
In contrast, truly unsupervised entity linkers should be able to operate in the \textit{absolute absence of annotated data}, with access to only a reference knowledge graph (KG) and a list of entity names, or aliases. Furthermore, neither the candidate generator nor the disambiguation technique (the two key modules of any typical entity linker) can make use of annotated data. Therefore, ``unsupervised'' disambiguation techniques that leverage labeled data to generate candidate entities \cite{pan2015unsupervised} are not applicable in our setting. Only very recently have researchers \cite{titov_unsup,Logeswaran2019ZeroShotEL} started to focus on EL systems that can operate in the absolute absence of annotated data. The motivation is well founded: there are some domains that link to their own specialized KGs, for which labeled data is not readily available, expensive to obtain, or scarce in the best case. This can be exemplified with domains such as law, science, or medicine. Moreover, companies may have their proprietary KGs where some entities are only meaningful with respect to the company. In all these cases, available labeled data cannot generalize to the corresponding specialized KGs.

\xhdr{Challenges} The majority of the existing methods \cite{milne,ganea,yamada_tacl,gupta2017entity,el1} in the EL literature are ill-suited for unsupervised EL as they rely heavily on annotated data. These methods typically leverage annotated data to (1) generate candidate entities, (2) use features (\eg the prior probability $P(e|m)$ of an entity $e$ given a mention $m$) derived from these annotations, (3) learn aligned word and entity embeddings enabling computation of similarity between an entity and the mention context, and (4) use a set of mentions and their corresponding gold entities to train supervised models. Thus, it is non-trivial and unclear how to adapt these methods to work in the absence of annotated data.

By the same token, one may not expect a large validation set of annotated data wherein to perform extensive hyperparameter tuning. Thus, novel approaches to EL without annotated data should either possess a small number of hyperparameters or be robust to hyperparameter tuning.

\xhdr{\ethemes} We propose \ethemes, a scalable approach for performing collective disambiguation in the absence of annotated data. Given a document with a number of mentions, each mention with a set of candidate entities that it can be linked to, and vector representations for all these candidate entities, \ethemes builds a matrix of the vector representations of all candidate entities within the document, and uses singular value decomposition to learn a subspace spanned by a number $k$ of components referred to as ``eigenthemes'', where $k$ is the only hyper-parameter specific to \ethemes. Note that each principal component of the learned subspace captures the topical relatedness among entities across a different ``latent'' facet, and thus, by keeping several principal components, \ethemes is equipped to deal with multi-topic text.

By design, \ethemes is suitable in settings where the gold entities within a document are topically related, in particular much more so than other subsets of candidate entities---a realistic assumption made by almost all existing works in the EL literature \cite{cucerzan2007large,yamada,xel3}, and for which we also provide a \emph{data-driven verification} in \S~\ref{sec:analysis}. 
By virtue of this assumption, one may infer that the mentions \emph{Michael Jordan} and \emph{Science} in the example from Fig.~\ref{fig:motivation} link to entities from the realm of \emph{science}. Moreover, as illustrated for toy data in Fig.~\ref{fig:motivation} and later verified for real data (\cf \S~\ref{sec:analysis}), gold entities tend to form a semantically dense subset of all candidate entities in the document and are therefore expected to lie in the subspace or in its proximity. 
\ethemes defines a similarity function that will output high values for entities whose embeddings lie in the proximity of the subspace. Eventually, these scores are used to perform collective disambiguation. 

\xhdr{Contributions}
\begin{itemize}[nolistsep,leftmargin=4mm]
	\item We propose \ethemes, a light-weight and fully unsupervised approach possessing the capability to incorporate external signals as weights for learning improved subspaces (\S~\ref{sec:approach}).
    \item We propose multiple strong baselines that compare favorably to (and sometimes even outperform) the current state of the art \titov \cite{titov_unsup}, and showcase the superiority of \ethemes over the considered methods on four benchmark datasets (\S~\ref{sec:exp}).
    \item For the first time, we provide a data-driven justification for the popular and long-standing assumption regarding the relatedness among gold entities mentioned within a document being higher than any other subset of candidate entities (\S~\ref{sec:analysis}).
\end{itemize}
\section{Related Work}
\label{sec:rwork}
\xhdr{Unsupervised EL} 
Recent works \cite{titov_unsup,Logeswaran2019ZeroShotEL} have spurred interest in EL with no or limited annotated data. \citet{titov_unsup} propose \titov, which uses distant supervision to compute a compatibility score between each of the candidate entities and the context of the mention. Entities are decomposed as sets of entity properties (types) and their representations are learned simultaneously with all other model parameters. The model is trained using a weak signal for scoring all entities in the (positive) set of candidates of a mention higher than all entities in a (negative) set of randomly sampled entities. 
Candidate entities are ranked based on the computed local compatibility scores. While the setting is identical to ours, the approaches are radically different. 
A downside of \titov \cite{titov_unsup} is that it requires validation of a large number ($15$) of hyperparameters, which becomes problematic if the performance relies on an extensive hyper-parameter tuning on a per-dataset basis. 
\citet{Logeswaran2019ZeroShotEL} train an entity linking model using an annotated dataset and apply it to another dataset where annotations are not available. 
Entities are only identified by (long) textual descriptions, and
the model relies strictly on language understanding to resolve the new entities. Distant supervision for EL has also been explored by \citet{el_distant}, however, their approach relies on features extracted from entity descriptions available from Wikipedia articles. We do not compare to \citet{Logeswaran2019ZeroShotEL} and \citet{el_distant}, as, unlike ours, their approach cannot be generalized to set-ups where such entity descriptions are not available.

\xhdr{Applications of low-rank subspaces} 
While low-rank subspaces have been employed in diverse application domains, \eg, face recognition \cite{turk1991eigenfaces}, link prediction \cite{west_eigenarticles}, distributional semantics \cite{word2vec_matfact, Mu2017RepresentingSA}, sense disambiguation \cite{Mu2016GeometryOP}, \etc, to the best of our knowledge, they have never been used for performing EL. We also emphasize our non-obvious use of low-rank subspaces: (1) we compute one decomposition per document, whereas most aforementioned applications compute a single corpus-wide decomposition, and (2) we incorporate weights.
\section{Problem and Notation}
\label{sec:problem}
We first formalize the task of interest. Let $D$ be a single document from a collection $\mathcal{D}$ of documents. 
Similar to most previous works in the EL literature, with few exceptions~\citep{sil2013re,Kolitsas2018EndtoEndNE}, we assume that mention spans (usually obtained by a named entity recognizer) are provided.
Also, let $\mathcal{M}_D = \{m_1, m_2, \dots, m_N\}$ be the set of $N$ mentions contained in $D$, and $\mathcal{E}$ be the set of entities in the reference KG $\mathcal{G}$. A low-dimensional representation (embedding) can be learned for each entity by applying node representation learning techniques such as \dw \cite{deepwalk} to the graph $\mathcal{G}$. The entity embeddings learned by these techniques are known to be meaningful with respect to the relatedness of the entities they represent \cite{Almasian2019WordEF}. The EL task consists in finding, for each mention $m \in \mathcal{M}_D$, the entity $e \in \mathcal{E}$ to which it refers. A key component of any EL system is the candidate generation system. Let $\mathcal{C}_T$ denote such a system, which, for a given mention $m$, retrieves at most $T$ entities---typically referred to as candidate entities. In the simplest case, the candidate generation system may retrieve all possible candidates (then $T=\infty$) \citep{yamada} for a given mention, but it is common to most of the works to retrieve a small subset of the most likely entities for that mention based on a certain sorting criterion. There is no consensus in the literature regarding a suitable value of $T$, and previous works have used varying values such as 7 \citep{ganea}, 20 \citep{xel4} or 30 \citep{Kolitsas2018EndtoEndNE}. For each mention $m \in \mathcal{M}_D$, the set of candidate entities is given by $\mathcal{C}_T(m) = \{e_{m}^1, e_{m}^2, \dots, e_{m}^T\}$, where $e_{m}^i$ indicates the $i^{th}$ candidate entity for the mention $m$. For the example introduced in \S~\ref{sec:intro}, the candidate entities for the mentions \textit{Michael Jordan} and \textit{Science} are portrayed in Fig.~\ref{fig:motivation}.

A second key component of any entity linker is the disambiguation module. This module selects an entity contained in the referent KG $\mathcal{G}$ for each mention of a document. The disambiguation process may be performed independently for each mention, or collectively for all mentions in a document. In the latter case, the fundamental assumption of all collective disambiguation algorithms is that the set of gold entities are topically related. 

\xhdr{Setting} Contrary to most of the existing work, where both the candidate generation and the disambiguation modules are dependent on annotated data, we focus on the setting where we only know the referent KG $\mathcal{G}$ and the mentions to be disambiguated, but have no annotated data that could generalize to the entities contained in $\mathcal{G}$.
\section{Entity Linking with \ethemes}
\label{sec:approach}
\ethemes learns a low-rank subspace from the ensemble of embeddings corresponding to all candidate entities for all mentions within a document (\S~\ref{subsec:ethemes}). We will see in \S~\ref{sec:weighted-eigenthemes} that the learning of the subspace may also be guided by different weights associated with the embeddings. The learned subspace is represented by what we refer to as ``eigenthemes''. The eigenthemes are components that are learned so as to decompose the ensemble of the entity embeddings as a linear combination of these components.

The vector--subspace similarity function (\S~\ref{sec:similarity-function}) takes the learned low-rank subspace and the vector representation of a candidate entity, and computes a similarity score as a weighted sum of similarities between the entity vector representation and each of the eigenthemes of the subspace. 
The similarity between the entity representation and an eigentheme indicates how much of the signal from the former can be projected into the latter, and it is weighted with a value that relates to the strength of the eigentheme in the entity embedding ensemble of the document. 
As a result, only those entities that lie in the learned subspace or in its proximity will have a high score. 
It is important to note that while the vector--subspace similarity function is applied to each mention's candidates independently, the learned subspace encompasses information from all mentions in a document. Therefore, the disambiguation performed by \ethemes is collective, as the subspace given by the eigenthemes is learned at a global level in the document.

Additional details (pseudocode, architectural overview, \etc) about \ethemes are presented in Appendix~\ref{app:approach}. 

\subsection{Subspace Learning}
\label{subsec:ethemes}
Let $\mathbf{E}$ be the $d$-dimensional embeddings for the entities $\mathcal{E}$ in the reference knowledge graph $\mathcal{G}$. Given a document $D$, we build the candidate space $CS$ by taking the union of the candidate entities given by $\mathcal{C}_T(m)$ for all the mentions  $m \in \mathcal{M}_D$ in the document. The $n_D \times d$ document embedding matrix $\mathbf{E}_D$ results from mapping all the $n_D$ entities in $CS$ to their corresponding embeddings in $\mathbf{E}$ and stacking them as rows. 

To learn the subspace $\mathcal{S}_D$ of $\mathbf{E}_D$, we use the Singular Value Decomposition (SVD), one of the most fundamental techniques for matrix factorization. The SVD of $\mathbf{E}_D$ decomposes each entity embedding in the matrix as a linear combination of left and right singular vectors, as well as singular values. The truncated SVD obtains a rank-$k$ ($k < \min(n_D,d)$) approximation $\mathbf{\tilde{E}}_D$ to $\mathbf{E}_D$ by just keeping the $k$ largest singular values and their associated left and right singular vectors. The approximated rank-$k$ matrix is
\begin{equation}
    \moveup
    \mathbf{\tilde{E}}_D = \mathbf{U}_k \mathbf{\Sigma}_k \mathbf{V}_k^{\top}.
    \label{eq:truncated-svd}
\end{equation}
where $\mathbf{U}_k$ and $\mathbf{V}_k$ are the $n_D \times k$ and $d \times k$ orthonormal matrices that preserve the $k$ left and right singular vectors respectively, and $\mathbf{\Sigma}_k$ is a $k \times k$ diagonal matrix of the $k$ largest singular values.

Each entity embedding is approximated as a linear combination---with coefficients given by the rows of $\mathbf{U}_k \mathbf{\Sigma}_k$---of the right singular vectors, \ie the columns of $\mathbf{V}_k$. The columns of $\mathbf{V}_k$ are the eigenthemes that form the subspace (hyperplane) to be used to perform collective disambiguation. The aforementioned coefficients are determined by $\mathbf{U}_k$, whose rows are entity-specific, and $\mathbf{\Sigma}_k$, which relates to the global strength of each eigentheme in $\mathbf{E}_D$. As discussed previously, the proposed similarity function to score candidate entities leverages both eigenthemes $\mathbf{V}_k$ and their strength $\mathbf{\Sigma}_k$.

\xhdr{Avoiding norm induced bias}
\citeauthor{eckart1936approximation} \cite{eckart1936approximation} proved that $\mathbf{\tilde{E}}_D$ is the solution to the following optimization problem.
\begin{equation}
    \nonumber
    \moveups
    {\arg\min}_{\mathbf{X}} ||\mathbf{E}_D - \mathbf{X}||^2_F \quad \text{subject to }\text{rank(\textbf{X}) = }k.
\end{equation}
That is, $\mathbf{\tilde{E}}_D$ is the rank-$k$ approximation to $\mathbf{E}_D$ that minimizes the Frobenius norm between both matrices. Thus, the truncated SVD is affected by the norm of the embeddings used to construct $\mathbf{E}_D$. It is for this reason that, prior to the learning of the subspace, each of the entity embeddings of the $\mathbf{E}_D$ matrix is normalized to have unit $L_2$ norm. Otherwise, the learning of the subspace would be driven by entities whose embeddings' norm is larger. Having normalized all the embeddings and chosen the number $k$ of components to be small, only the data points of the dense region of the embedding space will lie in the learned subspace or in its proximity.

\subsection{Incorporating Weights}
\label{sec:weighted-eigenthemes}

Let $\mathcal{W}$ be a weighting scheme\footnote{Weighting schemes are described in Appendix~\ref{app:weighting}.} for assigning weights to all the candidate entities in a document based on an external signal, which may capture a prior likelihood of an entity to be the gold entity. The weights provided by $\mathcal{W}$ can serve to guide the learning of the subspace. The weights associated with entities have an impact in the optimization problem solved by the SVD. Consequently, the low-rank decomposition prioritizes accurate approximation of the entity embeddings contained in the document embedding matrix based on both the density of the entities in the embedding space and their associated weights. 

We follow a simple approach to incorporate weights in the subspace learning by scaling each row of the document embedding matrix $\mathbf{E}_D$ with the weights given by an $n_D \times n_D$ diagonal matrix $\mathbf{W}_D$. Therefore, the subspace $\mathcal{S}_D$ is learned from the rank-$k$ approximation of $\mathbf{W}_D \mathbf{E}_D$. The values in $\mathbf{W}_D$ are obtained by applying the weighting scheme $\mathcal{W}$ to each of the $n_D$ candidate entities. The weighting scheme enriched subspace is then learned by performing an eigendecomposition of the weighted sums of squares and cross product (SSCP) matrix, which is formally stated as follows: 
\begin{equation}
    \moveups
    (\mathbf{W}_D \mathbf{E}_D)^{\top}(\mathbf{W}_D \mathbf{E}_D) = \mathbf{E}_D^\top \mathbf{W}_D^2 \mathbf{E}_D.
    \label{eq:weighted_sscp}
\end{equation}

The benefits of this extension are: (1) each candidate entity is enriched with some evidence about being a gold entity, and (2) the number of candidates per mention $T$ can be increased without having a considerable negative impact on the learned subspace, as unlikely candidate entities will be penalized by the weighting scheme. Empirically, weighted \ethemes obtains better performance and is more robust to the parameter $T$ than its vanilla unweighted version (detailed analysis in Appendix~\ref{app:weighting}).

\subsection{Similarity Function}
\label{sec:similarity-function}
The subspace $\mathcal{S}_D$, determined by the eigenthemes $\mathbf{V}_k$ and their strength $\mathbf{\Sigma}_k$, learned from all the mentions in a document is at the core of our similarity function, and allows to disambiguate every mention in the document. 
For each mention, we project the embeddings of the candidate entities into the subspace $\mathcal{S}_D$ and select the candidate with the largest norm. Candidate entities close to the subspace will score highly, while those orthogonal to the subspace will obtain a low score. 
Formally, for a mention $m$, the score of the $i^{th}$ candidate entity $e_m^i$ is computed as follows:
\begin{equation}
    \moveups
    \text{score}(e_m^i) = || \mathbf{e}_m^i \mathbf{V}_k \mathbf{\Sigma}_k||_2,
    \label{eq:similarity}
\end{equation}
where $\mathbf{e}_m^i$ is the embedding of the entity $e_m^i$. We observed that re-scaling the projection of the entity embedding into the eigenthemes with the corresponding strengths $\Sigma_k$ leads to better performance.
\section{Experiments}
\label{sec:exp}
All the resources (code, datasets, \etc) required to reproduce the experiments in this paper are available at~\url{https://github.com/epfl-dlab/eigenthemes}.

\subsection{Datasets}
\label{subsec:data}
We present results on real-world benchmark datasets (Table~\ref{tab:datasets}) that are popular in the entity linking literature. The considered datasets constitute a judicious mix of scale and domain types. For details about the datasets, please see Appendix~\ref{app:data}.

\xhdr{Candidate generation} We employ a simple and efficient approach to generate candidates, which is similar in design to the candidate generator used by \citet{titov_unsup}. Given a mention $m$, the entities that contain all the words from $m$ are considered candidate entities. For example, \textsc{Michael Jordan (basketball player)} and \textsc{Michael Jordan (computer scientist)} are candidates for the mention \textsc{Michael Jordan}, while \textsc{Michael Jackson} is not. Since the degree of an entity roughly captures its popularity, the candidate entities are sorted based on the degree of their corresponding vertices in the undirected version of the Wikidata KG. Although simple, the effectiveness and practicality of the candidate generator is corroborated by the high oracle recall (percentage of mentions where the true entity is present in the set of candidates) obtained across all the datasets presented in Table~\ref{tab:datasets}. It is important to note that the methodology of \ethemes is orthogonal to the candidate generator. While the latter could be improved (by modifying the string matching heuristic \cite{sil2012linking,candgen1}, or using word embeddings to match words in entity names and words in the mention, \etc), which will only improve the performance of any technique, it is beyond the scope of this work.

\begin{table}[t]
\caption{Datasets and their statistics. Mentions where the gold-entity was not contained in the candidate-set are marked `Not-found'. The `easy' and `hard' categorization of mentions is discussed in \S~\ref{subsec:data}.}
\moveups
\label{tab:datasets}
\centering
\resizebox{0.99\linewidth}{!}{
\begin{tabular}{c||c|c|c|c|c}
\hline
\multirow{2}{*}{\bf Dataset} & \multicolumn{3}{c|}{\bf \#Mentions} & \multirow{2}{*}{\bf \#Documents} & \multicolumn{1}{c}{\bf Oracle} \\
\cline{2-4}
& \bf \#Easy & \bf \#Hard & \bf \#Not-found & & \multicolumn{1}{c}{\bf Recall} \\
\hline
\textbf{CoNLL-Val} & 2645 & 1542 & 590 & 216 & 87.7\% \\
\textbf{CoNLL-Test} & 2555 & 1417 & 506 & 231 & 88.7\% \\
\textbf{WNED-Wiki} & 2731 & 3025 & 938 & 318 & 85.9\% \\
\textbf{WNED-Clueweb} & 4667 & 4711 & 1662 & 320 & 84.9\% \\
\textbf{Wikilinks-Random} & 26373 & 17397 & 10729 & 2336 & 80.3\%\\
\hline
\end{tabular}
}
\moveup
\moveup
\moveup
\end{table}

\begin{table*}[t]
\moveup
\caption{EL quality measured using precision@1 and MRR on the CoNLL-Test dataset. The best performance is shown in \textbf{bold}. For \titov and \wtitov, we perform $5$ independent runs and report the mean and standard deviation. All other techniques are deterministic. Note that the parameter $T$ was fixed to $20$ candidates per mention, and thus, `\#Hard' mentions reduces from $1417$ to $1136$ while `\#Not-found' increases from $506$ to $787$.}
\label{tab:conll_results}
\moveups
\centering
\begin{threeparttable}
\resizebox{0.99\linewidth}{!}{
\begin{tabular}{llcccccc}
\toprule
\multirow{2}{*}{\bf Category} & \multirow{2}{*}{\bf Method} & \multicolumn{3}{c}{\bf Precision@1} & \multicolumn{3}{c}{\bf MRR} \\
\cmidrule(lr){3-5}
\cmidrule(lr){6-8}
&  & \bf Overall\tnote{\#} & \bf Easy & \bf Hard & \bf Overall\tnote{\#} & \bf Easy & \bf Hard \\
\midrule
Existent & \nm \cite{nm} & 0.412 & 0.645 & 0.174 & 0.415 & 0.645 & 0.184 \\
Existent &  \titov (SoTA) \cite{titov_unsup} & 0.451 $\pm$ 0.019 & 0.700 $\pm$ 0.032 & 0.187 $\pm$ 0.006 & 0.539 $\pm$ 0.017 & 0.777 $\pm$ 0.029 & 0.353 $\pm$ 0.005 \\
\midrule
Proposed & \lctxt & 0.296 & 0.420 & 0.223 & 0.401 & 0.537 & 0.374 \\
Proposed & \gctxt & 0.303 & 0.403 & 0.289 & 0.423 & 0.542 & 0.448 \\
Proposed & \degree & 0.571 & \bf 1.0\tnote{\dag} & 0.0 & 0.649 & \bf 1.0\tnote{\dag} & 0.312 \\
\midrule
Proposed & \avg & 0.488 & 0.658 & 0.445 & 0.593 & 0.756 & 0.636 \\
Proposed & \wtitov & 0.499 $\pm$ 0.022 & 0.778 $\pm$ 0.037 & 0.217 $\pm$ 0.008 & 0.592 $\pm$ 0.018 & 0.853 $\pm$ 0.030 & 0.415 $\pm$ 0.007 \\
\midrule
Proposed & \eshort & \bf 0.617\tnote{\dag} & 0.858 & \bf 0.500\tnote{\dag} & \bf 0.690\tnote{\dag} & 0.910 & \bf 0.674\tnote{\dag} \\
\midrule
- & Ceiling & 0.824 & 1.0 & 1.0 & 0.824 & 1.0 & 1.0 \\
\bottomrule
\end{tabular}
}
\begin{tablenotes}
\small \item[\dag] Indicates statistical significance ($p<0.01$) between the best and the second-best method using the Student's paired $t$-test.
\small \item[\#] Overall is computed by considering all mentions (including Not-found in addition to Easy and Hard).
\end{tablenotes}
\end{threeparttable}
\moveup
\moveup
\moveups
\end{table*}

\xhdr{Preprocessing} We consider Wikidata 
as our referent KB. The gold entity annotations for mentions available as Wikipedia page ids in all the aforementioned datasets were appropriately mapped to their corresponding Wikidata identifiers. To ensure that our empirical analyses and the corresponding conclusions are representative of various real-world scenarios requiring entity linking, we, similar to \citet{xel3} and \citet{wned}, introduce the \emph{`easy'} and \emph{`hard'} categorization for mentions in all the datasets. A mention is termed as `easy' if the first candidate entity (in the list of sorted candidates returned by the candidate generator) is the `true' entity, and `hard' otherwise. 

\subsection{Methods Benchmarked}
\label{subsec:baselines}

\xhdr{Existing methods} To the best of our knowledge, surface name matching (\nm) and \titov are the only methods tried in previous work.

\noindent $\bullet$ \textbf{\nm \cite{nm}:} For each mention $m$, \nm retrieves all Wikidata entities whose names match exactly with the mention string. In the event of multiple matching entities, we choose the entity with the highest KG degree as the prediction for mention $m$.

\noindent $\bullet$ \textbf{\titov \cite{titov_unsup}:} The current state of the art (SoTA), which is trained using the New York Times dataset provided by \citeauthor{titov_unsup}.

\xhdr{Newly introduced baselines} We introduce five creative and simple, yet effective baselines.

\noindent $\bullet$ \textbf{\lctxt:} computes a representation of the local context of a mention as the average of its surrounding words' embeddings. For each mention $m$, candidate entities are ranked based on their semantic similarity with the context, and the one with the highest similarity is chosen as the prediction for $m$. Entity representations are computed from their textual descriptions (details in Appendix~\ref{app:entity_embed}). We set the context window size to $5$\footnote{Other window sizes were evaluated, however, the chosen value $5$ resulted in the best downstream EL performance.}.

\noindent $\bullet$ \textbf{\gctxt:} follows the exact same procedure as \lctxt, with the only difference being the size of the context window. Following convention in the literature \cite{yamada}, we use all the nouns in a document to obtain the global context representation.

\noindent $\bullet$ \textbf{\degree:} a natural baseline (courtesy of our candidate generator) for performing EL without annotated data. For each mention $m$, entity degree is used to obtain a ranking of the candidate entities, and the one with the highest degree is chosen as the prediction for $m$.

\noindent $\bullet$ \textbf{Average (\avg):} constructs a representation of $\mathbf{E}_D$ as a $d$-dimensional vector by computing the average of the rows of $\mathbf{E}_D$. Each candidate entity is scored by computing the cosine similarity between its embedding and the \avg based representation.

\noindent $\bullet$ \textbf{\wtitov:} extends the current SoTA \titov by incorporating the weights used by \eshort into its compatibility scoring function.

\xhdr{\ethemes (\eshort)} Our proposed solution. 

Unless stated otherwise, weights are employed for \avg and \eshort. Note that we opt against purely supervised baselines as running them correctly in the ``true'' unsupervised setting is inconceivable (\cf challenges in \S~\ref{sec:intro}).

\begin{figure*}[t]
\moveup
\moveup
\centering
    \includegraphics[width=0.9\linewidth]{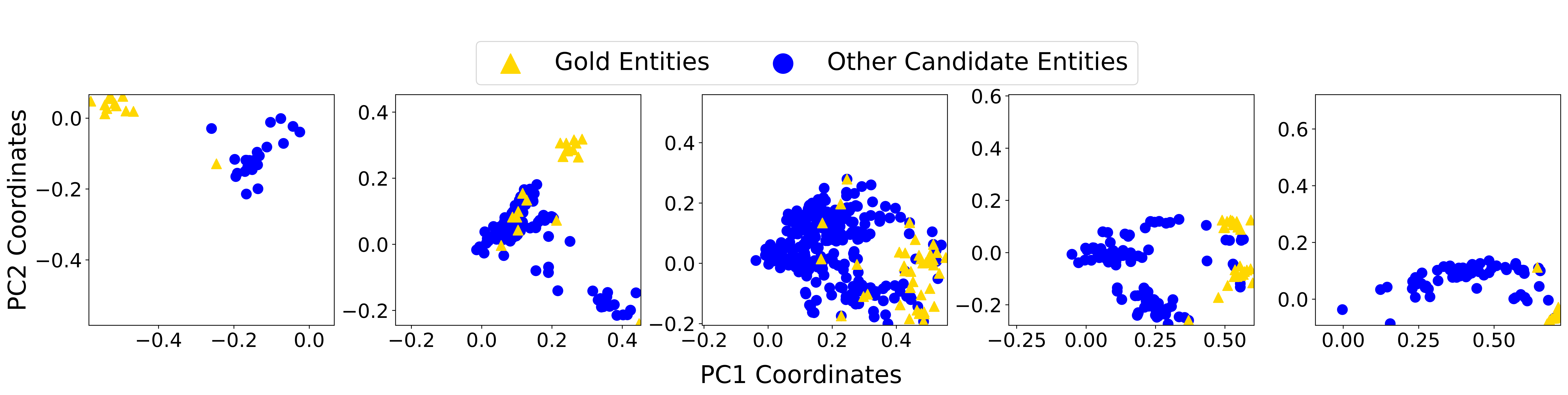}
    \figcaption{Scatter plots for $5$ documents sampled randomly from the CoNLL-Test dataset. For each document, we project all the candidate entities on the first two components (PC1 and PC2) of the subspace learned by \eshort.}
\label{fig:analysis_subspace_pc}
\end{figure*}

\subsection{Setup}
\label{subsec:setup}
\xhdr{Embeddings} For word embeddings, we use the publicly available \cite{word2vec_google} 300-dimensional vector representation of words obtained by training \wv \cite{word2vec} on the Google News dataset. The 128-dimensional entity embeddings were obtained by training \dw \cite{deepwalk} on the Wikidata knowledge graph. 

\xhdr{Weighting scheme} The candidate entities of a mention are weighted using their reciprocal rank, where the ranking is induced by the candidate generator \cite{titov_unsup}. 

\xhdr{Parameters} The maximum number of candidates per mention $T$ is fixed to $20$. Note that restricting $T$ also results in a reduction in the oracle recall. We fix the number $k$ of components for constructing the subspace representation using \eshort to $10$. 

\xhdr{Evaluation metrics} We use (1) precision@1, and (2) mean reciprocal rank (MRR) to evaluate the quality of the methods benchmarked in this study. Following convention in the literature, we compute \emph{Micro} aggregates of the metrics over all mentions. 

Additional details about the experimental setup (effect of different embedding methods, weighting schemes, the number of candidates $T$, \etc), and hyperparameter tuning (effect of $k$) are present in Appendix~\ref{app:setup} and~\ref{app:tuning}, respectively.

\subsection{Results: CoNLL}
\label{sec:results_conll}
We assess the efficacy of \eshort by comparing its quality measured using precision@1 and MRR with the considered methods on the CoNLL-Test dataset. The results are presented in Table~\ref{tab:conll_results}. Note that Ceiling corresponds to the oracle recall of the candidate generator and provides an upper bound on precision@1 and MRR. 

\xhdr{Overall performance}
It is evident that \eshort achieves the best overall quality and significantly outperforms all the considered methods. The key highlights are as follows: (1) \eshort obtains an improvement of $15$ percentage points over the existing SoTA \titov, (2) \avg, a hard-to-beat baseline \cite{avg_sentence, simple_gcn} and a natural competitor of \eshort, is around $12$ percentage points inferior to \eshort, which substantiates our intuition (\cf \S~\ref{sec:intro} and~\ref{sec:approach}) about the superiority of subspaces in capturing global topical relatedness among entities in a document, and (3) \degree, a simple baseline that we introduce in this work, substantially outperforms all the other considered methods---even the existing SoTA, \titov---and is second only to \eshort, which remains about $5$ percentage points and statistically significantly better (p-value $<0.01$) than \degree.

\xhdr{Performance on `easy' and `hard' mentions}
The key finding is that the closest competitors---\wtitov and \degree---of \eshort lack robustness to the variation in mention-types, which is substantiated by the huge disparity of EL performance measured using precision@1 (with similar trends observed for MRR) for both \wtitov ($78\%$ for easy vs. $22\%$ for hard mentions) and \degree ($100\%$ for easy vs. $0\%$ for hard mentions). This result highlights the key limitation of \wtitov and \degree that they cannot address challenging scenarios of EL. A detailed analysis about the potential effects of this limitation is performed in \S~\ref{sec:analysis}, while a discussion on the potential reasons behind the existence of this disparity in performance is presented in Appendix~\ref{app:disparity_easy_hard}.

\section{\textbf{Analysis}}
\label{sec:analysis}
In this section, we perform a post-mortem analysis on a plethora of aspects impacting the downstream EL performance of the considered methods measured using precision@1. Results for MRR show similar trends and are therefore omitted.

\xhdrNoPeriod{Do learned subspaces capture the relatedness among gold entities?}
Fig.~\ref{fig:analysis_subspace_pc} shows that gold entities tend to form tight clusters when projected on the first two components of the subspace learned by \eshort. This result provides a data-driven justification regarding the existence of relatedness among gold entities mentioned within a document---a key assumption used in the design of almost every method in the EL literature. Moreover, it also showcases the ability of the subspaces learned by \eshort to capture such relatedness.

\xhdrNoPeriod{Do gold entities lie closer to the learned subspaces when compared to other candidate entities?}
The disambiguation module of \eshort relies on the gold entities of a document being closer to the learned subspace than other candidate entities. While the strong EL performance of \eshort portrayed in Table~\ref{tab:conll_results} provides substantive evidence in favor of the aforementioned property, we conduct a more direct assessment, which is described as follows. For every mention in a document, we project all candidate entities onto the subspace learned by \eshort and compare the score (Eq.~\ref{eq:similarity}) of the gold entity to the average score (Eq.~\ref{eq:similarity}) of the non-gold entities, finding that the gold entity’s score $G$ is \emph{statistically significantly higher} ($81$\% on average) than the score $N$ of the average non-gold entity: $\frac{G-N}{N} = 0.81$, with bootstrapped $95$\% confidence interval $[0.78, 0.84]$. This finding convincingly substantiates the fact that incorrect candidate entities \emph{rarely} exhibit stronger topical relatedness than gold entities. In the event that incorrect candidates also possess strong relatedness, \eshort (by design) can differentiate between the true and spurious relatedness by leveraging weighting schemes.

\xhdrNoPeriod{Does the distribution of mention-types affect EL performance?}
Inspired by the \emph{cloze} test \cite{cloze} used to assess language learning capability of individuals \cite{mutilation}, we perform a \emph{mutilation analysis} to assess the impact of the distribution of mention-types on the downstream EL performance. The test is carried out by removing varying number of `easy' mentions from the CoNLL-Test dataset, thereby increasing the fraction of `hard' (ambiguous) mentions, and thus, can be deemed analogous to adding noise in the dataset. Specifically, we generate $11$ different dataset versions by subsampling (uniformly at random) varying fraction (between $1$ and $0$ in decrements of $0.1$) of easy mentions while retaining all the hard mentions, and measure the `overall' EL performance in each version. We run this experiment $10$ times and report the mean performance. Being very small, the standard deviations were omitted to avoid clutter.

\begin{figure}[t]
\moveup
\moveups
\centering
    \subfloat[Mutilation Analysis]
    {
        \scalebox{0.46}{
            \includegraphics[width=\linewidth]{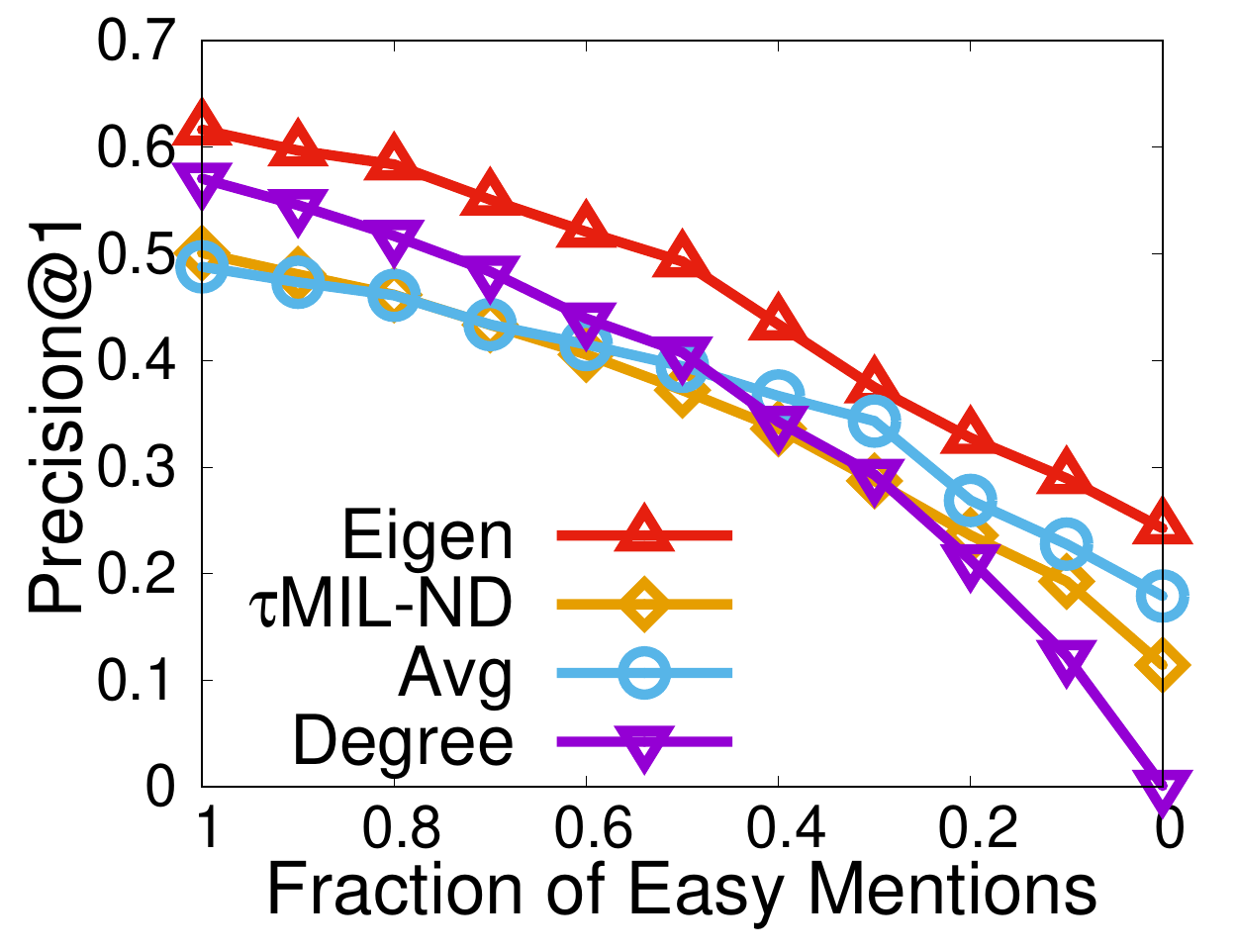}
        }
        \label{fig:mutilation_precision}
    }
    \subfloat[Out-of-domain Datasets]
    {
        \scalebox{0.46}{
            \includegraphics[width=\linewidth]{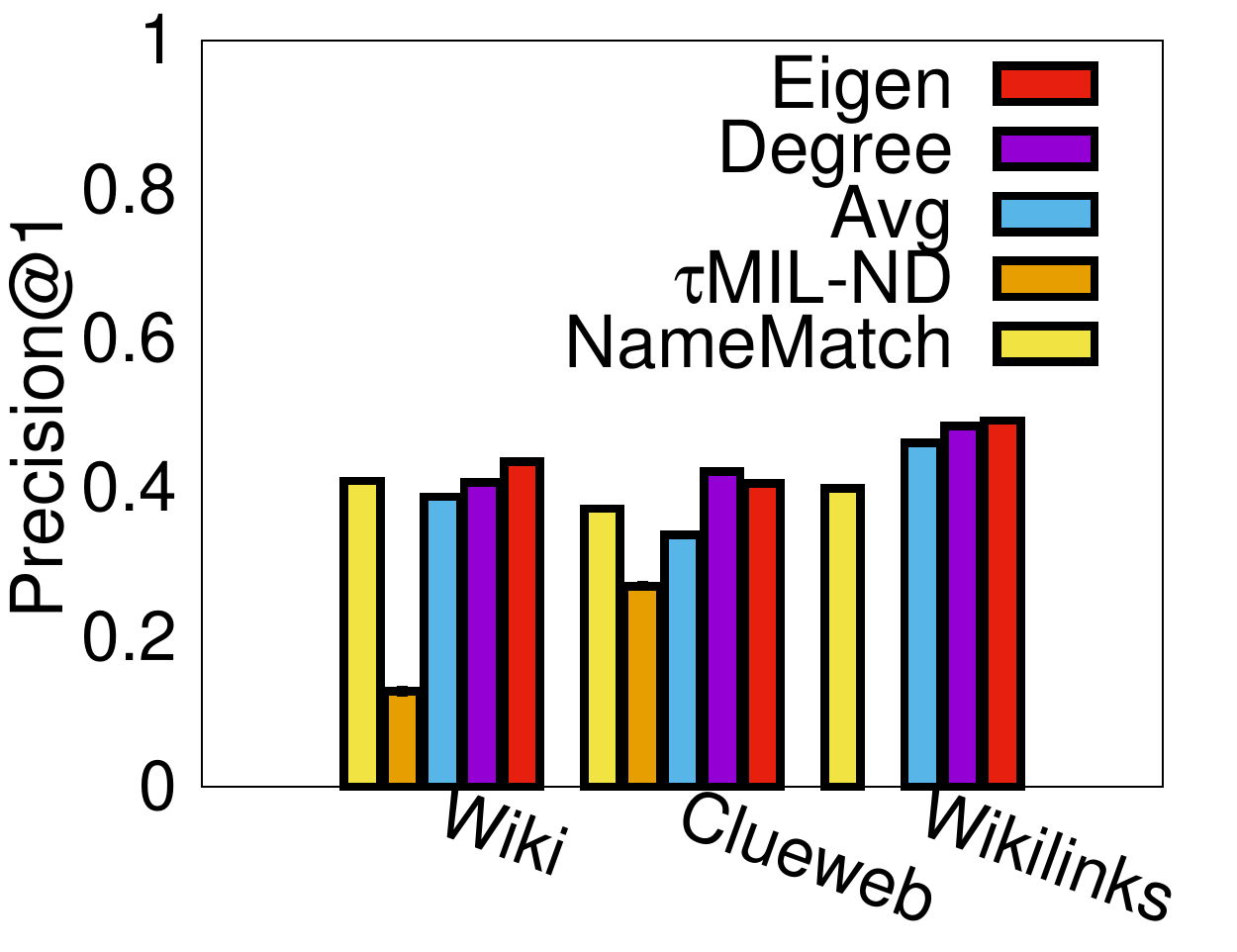}
        }
        \label{fig:out_of_domain_precision}
    }
    \figcaption{Analyzing the (a) effect of mention-type distribution, and (b) performance on out-of-domain datasets using precision@1.}
\label{fig:analysis_precision}
\moveups
\end{figure}

Fig.~\ref{fig:mutilation_precision} portrays that the overall EL performance of all the techniques deteriorates with the reduction in the fraction of easy mentions, which is natural. Interestingly, the deterioration observed for the \degree baseline is much more profound---precision@1 plummets from $0.57$ (fraction-easy = $1$) to $0$ (fraction-easy = $0$)---when compared to all other techniques, and it gets demoted from being the second best technique to the worst. Note that even \avg starts outperforming \degree when fraction of easy mentions are less than $0.5$. On the contrary, \eshort consistently outperforms all other techniques in all the subsampled versions.

This experiment establishes two important points. (1) It shows that the performance of \degree is \emph{deceptively high}. More importantly, it unveils a key limitation of \degree, \ie, its inability to address challenging scenarios of EL where the fraction of easy mentions is low. (2) It bolsters the superiority of \eshort over other techniques by showcasing its ability to reliably and consistently perform well, even in challenging scenarios.

\xhdr{Robustness: Out-of-domain Datasets}
It is standard practice in the EL literature \cite{ganea,el1} to evaluate the robustness of the models trained on the CoNLL-train dataset in out-of-domain datasets without any (re)validation of hyperparameters. In the same vein, the experiments on all the other datasets were performed using the hyperparameters tuned on the CoNLL-Val dataset. Specifically, we assess the robustness of the techniques on the recently introduced WNED-Wiki and WNED-Clueweb datasets, and the Wikilinks-Random dataset of English Wikipedia tables. The results are presented in Fig.~\ref{fig:out_of_domain_precision}. Note that \titov cannot perform EL in the Wikilinks-Random dataset owing to the absence of sequential textual content, and therefore, the corresponding bar in the plot is non-existent.

It is worth noting that \titov (the existing SoTA) performs the worst in the out-of-domain scenario, whereas all other techniques exhibit robustness. Hence, this experiment simply unveils a limitation of \titov, \ie, its lack of robustness, which is a noteworthy finding. Consequently, the experiment also establishes two important properties of \eshort: (1) robustness to hyperparameter tuning, and (2) adaptability to newer domains.
\section{Discussion}
\label{sec:discussion}

\begin{figure}[t]
    \moveups
    \centering
    \includegraphics[width=0.65\linewidth]{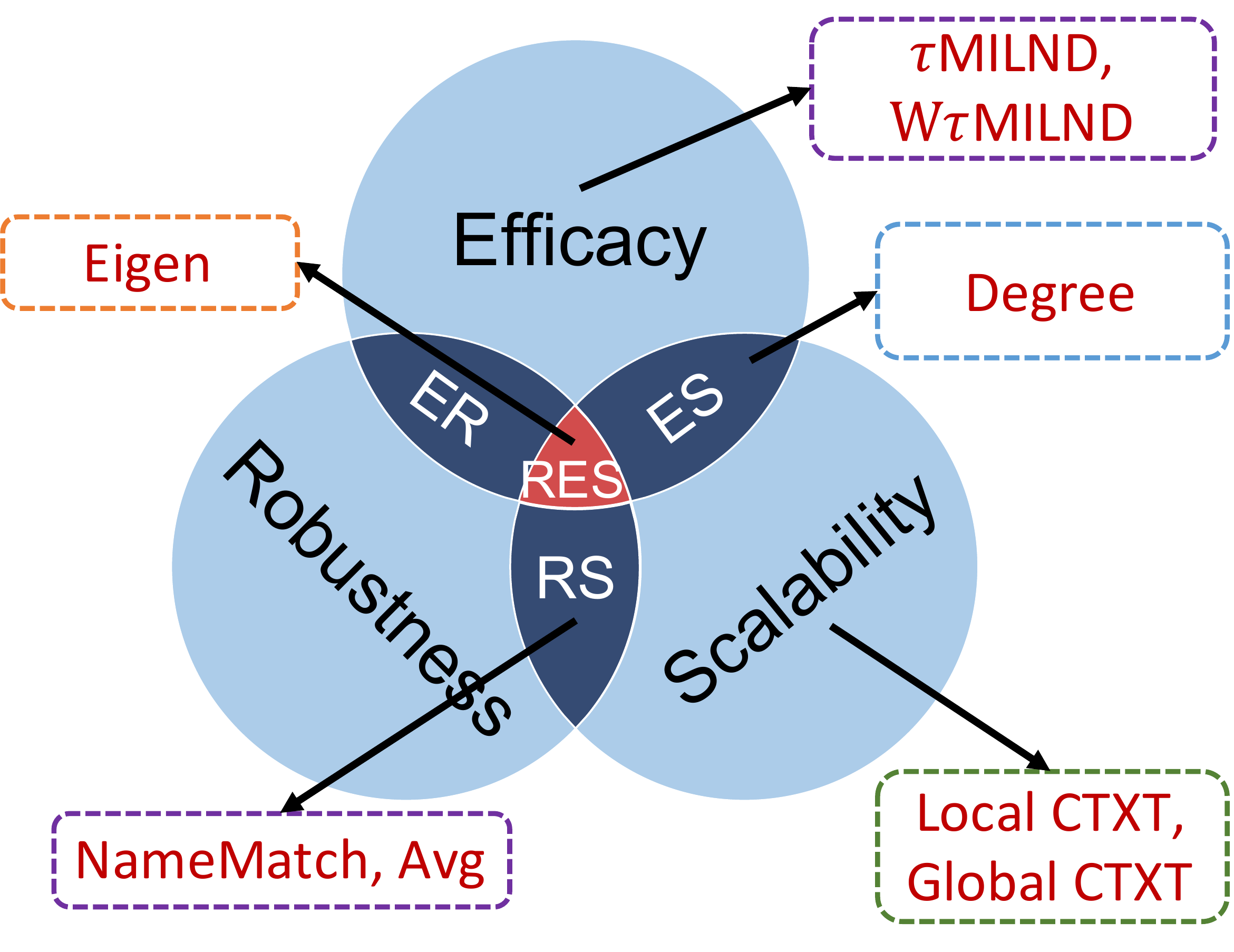}
    \figcaption{A qualitative summary of the spectrum of unsupervised EL techniques based on their strengths.}
    \label{fig:qualitative_summary}
\end{figure}

\xhdr{Summary of Results}
A strong entity linking method for scenarios with no annotated data should stand strong on three pillars: \underline{efficacy} towards entity disambiguation, \underline{scalability} to large datasets, and \underline{robustness} to hyperparameter tuning and mention-type distributions. We performed an extensive comparison of \eshort with $7$ techniques across each of the aforementioned features, and while a quantitative analysis has already been performed, we provide a visual summary in Fig.~\ref{fig:qualitative_summary}.

\eshort (classified as ``RES'') stands strong on all three pillars. (1) It is robust to the presence of noise in the data; provides simplicity of hyperparameter tuning as it possesses just one hyperparameter, \viz, the number of components to construct the subspace; and also portrays robustness to the hyperparameter tuning step. (2) Its light-weight nature provides the ability to gracefully scale to Web-scale datasets. Empirically, \eshort requires around $2$--$15$ minutes, while \titov \cite{titov_unsup} requires around $200$--$220$ minutes to perform entity linking for the considered datasets. Thus, our approach is approximately $10$ to $100$ times more efficient. (3) It portrays the best efficacy across all metrics, settings, and datasets. While \degree (classified as ``ES'') is a promising technique, its lack of robustness to the varying easy-hard proportion of queries in the datasets serves as a concerning disadvantage. As argued in the ``challenges'' paragraph of \S~\ref{sec:intro}, robustness is a desired characteristic in a setting where no annotated data is available.

Some of the additional advantages of our method include: (1) language independence owing to the reliance on entity embeddings alone, and (2) capability to improve existing methods in settings where annotated data is available (Appendix~\ref{app:supervised}). In summary, \ethemes provides an effective, efficient, and scalable solution to the entity linking problem in the absence of annotated data.

\xhdr{\ethemes \vs Clustering}
While Fig.~\ref{fig:analysis_subspace_pc} shows that gold entities tend to be clustered together in the eigenspace, producing internally-tight, mutually-far clusters (as obtained by, say, k-means \cite{kmeans1,kmeans2}) is neither the optimization objective nor a requirement for \ethemes to work. Moreover, using clustering to disambiguate gold entities from other candidate entities is easier said than done, as even with a perfect clustering where gold entities lie in a single coherent cluster, it is unclear how to identify which cluster contains these gold entities.

\xhdr{Domain-specific KGs}
Domain-specific KGs---associated with text corpora in niche domains such as cinema, law, medicine, or science---are usually sparser than Web-scale KGs such as Wikidata, which could lead to low-quality entity embeddings \cite{sparse_kg_low_quality}, thereby directly affecting the performance of \ethemes. To this end, we compare the information density (\#relationships-per-entity) \cite{sparse_kg_low_quality} of the Wikidata KG variant (Appendix~\ref{app:preprocess}) used in the current work to learn entity embeddings, with a plethora of domain-specific and Web-scale KGs. Specifically, the information density of our Wikidata KG (5.5) is similar to that of domain-specific KGs, namely: IMDb (4.5) \cite{imdb_data} and SNOMED (7.1) \cite{snomed}, and much smaller than Web-scale KGs, namely: DBpedia (26) \cite{dbpedia}, Freebase (16) \cite{freebase}, and Wikidata (12.5) \cite{wikidata}. Moving beyond aggregate statistics, even the degree distribution of our Wikidata KG is similar to that of the aforementioned domain-specific KGs. These statistics indicate that \ethemes can readily perform EL in domains with sparse KGs.

\xhdr{Unobserved Entities}
In its current state, \ethemes cannot perform EL for KG entities that were not observed during training of entity embeddings. Inductive learning \cite{graphsage} of entity embeddings serves as a potential fix, however, such extensions constitute as future work.

\section{Conclusions}
\label{sec:conclusions}

In  this  paper,  we  addressed  the  problem  of  EL in the absence of annotated data with a light-weight method, \ethemes, that relies solely on the availability of entity names and a referent KB. 
Experiments on benchmark datasets portrayed the effectiveness of our proposed approach.
In the future, our aim is to validate the effectiveness of our approach in performing multi-lingual entity linking with a special focus on low-resource languages.

\section*{Acknowledgements}
We thank Andreas Spitz, Maxime Peyrard, Sachin Kumar, and Sainyam Galhotra for insightful discussions. This project was partly funded by the Swiss National Science Foundation (grant 200021\_185043), the European Union (TAILOR, grant 952215), and the Microsoft Swiss Joint Research Center. We also gratefully acknowledge generous gifts from Facebook and Google.

\pagebreak
\section*{Broader Impact}
Entity linking is a broad problem with diverse application areas ranging from natural language processing to web data management. A less popular yet important application of entity linkers is their ability to act as an enabling technology for improving the navigability of networks such as Wikipedia, which depends heavily on the amount and diversity of hyperlinks joining pairs of articles. Moreover, they are also critical in facilitating users and machines to obtain a better understanding of text corpora by having ambiguous pieces of text linked to their corresponding unambiguous concepts. Entity linking could be deemed as a solved problem in scenarios where annotated data is available, however, there is a scarcity of methods capable of performing entity linking without access to annotated data. Consequently, designing effective and efficient solutions for entity linking without annotated data is an important avenue, and our current work is a step in that direction. The core contribution of this paper is a light-weight and language agnostic approach to unsupervised entity linking. While we are not the first to formulate the unsupervised entity linking problem, we are the first to propose a solution capable of operating on Web-scale datasets, which is a fundamental requirement for practical entity linkers. Moreover, we are also the first to introduce a suite of simple and intuitive, yet effective baselines that would serve as strong benchmarks, thereby enabling researchers to conduct follow-up research in this nascent but important research area.

\bibliographystyle{acl_natbib}
\balance
\bibliography{el}

\section*{Appendix}
\appendix
\section{Additional Related Work}
\label{app:rwork}

\citet{sil2012linking} address a related task in a setting similar to ours. Given a mention, they generate a restricted number of candidate entities, which very often contain the gold entity, by exploiting associated structured information. Very recently, \citet{medtype} address the same issue for medical entity linking by pruning irrelevant candidate entities based on the prediction of semantic type of a mention. We believe that these works are complementary to \ethemes and could bring important benefits to disambiguation techniques in the absence of annotated data.

\section{Additional Details about \ethemes}
\label{app:approach}
\begin{figure*}[t]
    \moveup
    \begin{minipage}{.56\textwidth}
        \scalebox{1}{
            \includegraphics[trim=0cm 0cm 0cm 0.7cm, width=\linewidth]{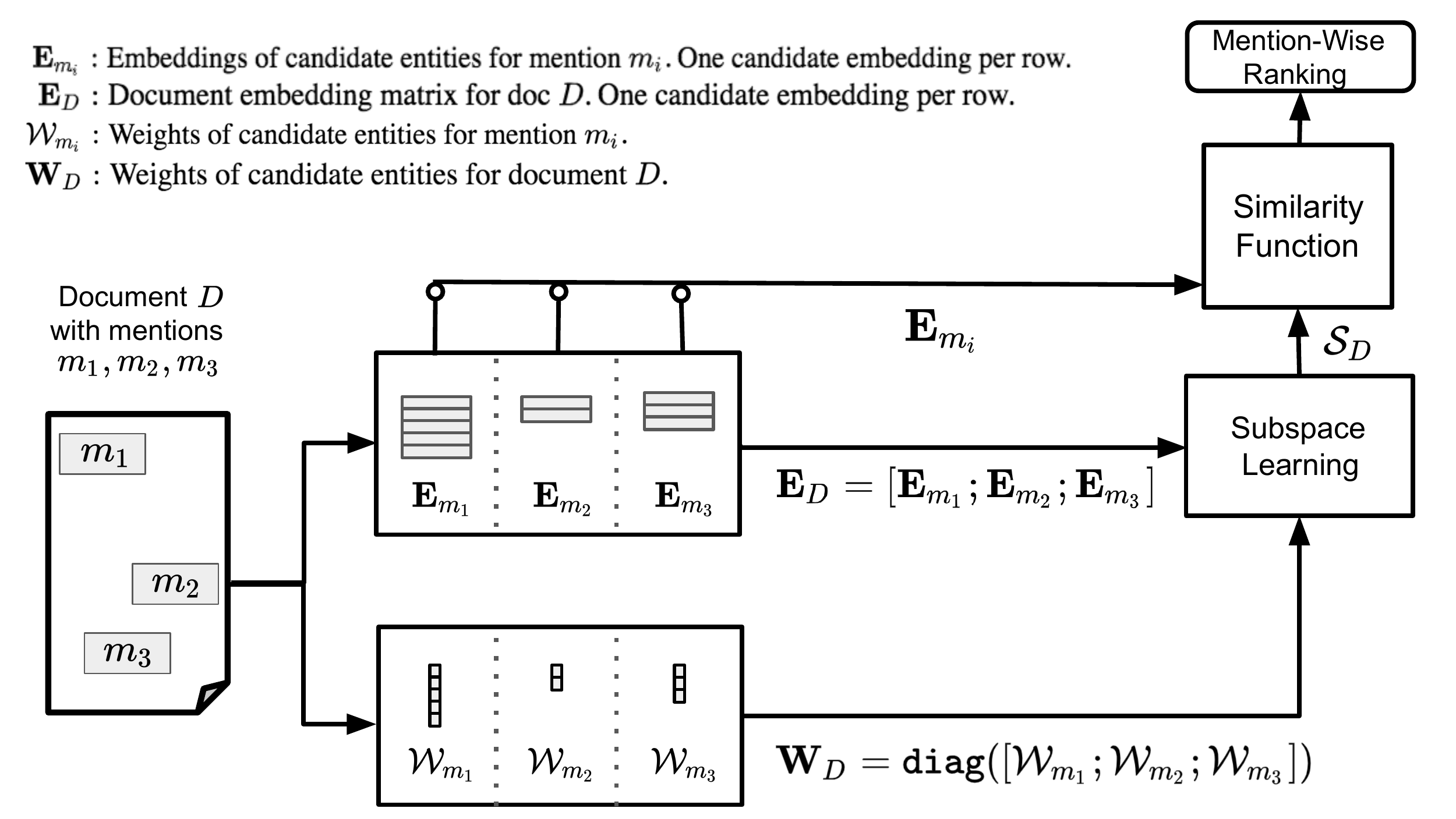}
            \label{fig:diagram}
        }
    \end{minipage}
    \hfill
    \begin{minipage}[t]{.4\textwidth}
            \vspace{-2.5cm}
            {\scriptsize
            \begin{algorithmic} [1]
        \REQUIRE A document $D$ which contains a set of mentions $\mathcal{M}_D$, a candidate generation system $\mathcal{C}_T$, a weighting scheme $\mathcal{W}$, $d$-dimensional entity embeddings $\mathbf{E}$, and the number of components $k$ to be learned by the subspace.
        \ENSURE Assignment of entities to each mention $m \in \mathcal{M}_D$
        \STATE Candidate space $CS \leftarrow \{\mathcal{C}_T(m) \ | \ m \in \mathcal{M}_D\}$
        \STATE $\mathbf{E}_D \leftarrow \text{Embeddings taken from $\mathbf{E}$ for entities in }CS$
        \STATE $\mathbf{W}_D \leftarrow \text{Weights taken from $\mathcal{W}$ for entities in }CS$
        \STATE Subspace $\mathcal{S}_D \leftarrow \text{Subspace\_Learning}(\mathbf{W}_D, \mathbf{E}_D, k)$
        \STATE Mention2Entity = \{\}
        \FORALL {$m \in \mathcal{M}_D$}
            \STATE score = \{\}
            \FORALL {$e \in \mathcal{C}_T(m)$}
                \STATE score[$e$] $\leftarrow$ Compute similarity between $\mathbf{e}$ and $\mathcal{S}_D$
            \ENDFOR
            \STATE Mention2Entity[$m$] = ${\arg \max}_{e'}$ score[$e'$]
        \ENDFOR
        \RETURN Mention2Entity
        \end{algorithmic}
        }
    \end{minipage}
    \figcaption{\label{fig:eigenthemes} (Left) Pipeline of \ethemes. (Right) Its pseudocode. $[\mathbf{a}; \mathbf{b}]$ is row-wise concatenation of $\mathbf{a}$ and $\mathbf{b}$.}
\end{figure*}

The pipeline of \ethemes and its pseudocode are depicted in Fig.~\ref{fig:eigenthemes}.

\xhdr{Discussion: The role of $T$ and $k$ in subspace learning} 
As stated in Sec.~\ref{sec:approach}, the eigenthemes are components that are learned so as to decompose the ensemble of the entity embeddings as a linear combination of these components. If the number $k$ of eigenthemes is chosen to be small, these components will constitute a good basis to approximate the dense region of the document embedding matrix. From the fundamental assumption of the existence of topical relatedness across the gold entities in a document and that such relatedness is captured by their corresponding embeddings \cite{Almasian2019WordEF}, the gold entities will form a dense region and, consequently, will define the subspace. However, this is only possible if there is no other subset of candidate entities whose relatedness is larger than that of the set of gold entities. The latter point relates to the hyperparameter $T$, which controls the maximum number of candidate entities per mention. A low value of $T$ reduces the possibility of having another subset of entities with larger relatedness as well as the recall---the number of times that the gold entity is contained in the set of candidate entities. A large value of $T$ increases both. In general, the larger the number of candidate entities per mention, the more the learned subspace will be affected by embeddings other than those from the gold entities.

\section{Datasets}
\label{app:data}

\subsection{\textbf{Detailed Dataset Description}}

\noindent $\bullet$ The \textbf{AIDA-CoNLL} dataset \cite{hoffart} is one of the most popular datasets in the ELliterature. It is based on the CoNLL 2003 shared task \cite{conll03_ner_task} and contains high quality manual annotations for mention strings linking them to their target named entities. The dataset consists of three parts: training, validation, and test, but, we use only the validation (CoNLL-Val) and test (CoNLL-Test) sets owing to the fully unsupervised nature of our approach.

\noindent $\bullet$ \textbf{WNED-Wiki} and \textbf{WNED-Clueweb} are recently introduced benchmark datasets by \citeauthor{wned} \cite{wned} aimed at reducing the bias in the AIDA-CoNLL dataset towards popular entities. Specifically, these datasets are generated by uniformly sampling mentions with different levels of difficulty (determined by the prior scores) from the `2013-06-06' dump of English Wikipedia and Freebase annotations of the 2012 Clueweb corpora (FACC1) \cite{facc} respectively.

\noindent $\bullet$ The \textbf{Wikilinks-Random} dataset \cite{bhagavatula2015tabel, limaye_tabel} consists of tables extracted from the English Wikipedia with mentions and their corresponding links to Wikipedia pages. Note that a key difference between this and the other datasets is the scarcity of textual content in the table data.

\subsection{\textbf{Dataset Preprocessing Details}}
\label{app:preprocess}

\xhdr{Referent KB} Wikidata \cite{wikidata} is considered as the referent KB. We use the n-triples format of the Wikidata dump\footnote{The `latest-truthy.nt.gz' file at \url{https://dumps.wikimedia.org/wikidatawiki/entities/}} downloaded on May 3, 2019. Since the gold entity annotations for mentions in all the datasets benchmarked in this study are provided as either page-titles or page-ids of the English Wikipedia, we restrict our entity universe to the ones having links to English Wikipedia resulting in about 3.7M Wikidata entities. The mapping between Wikipedia and Wikidata entities was extracted using the Wikimapper tool \cite{wikimapper} from the `page\_props.sql.gz' file available in the `2019-05-20' dump\footnote{\url{https://dumps.wikimedia.org/enwiki/}} of English Wikipedia. Eventually, each entity is denoted by its unique Wikidata identifier (QID). For example, the entity \textsc{Michael Jordan (computer scientist)} is denoted by his Wikidata QID: \textsc{Q3308285}. Note that $1$, $7$, $12$, and $6903$ mentions were ignored from the CoNLL-Val, WNED-Wiki, WNED-Clueweb, and Wikilinks-Random datasets respectively as their ground truth Wikipedia entities could not be mapped to their corresponding Wikidata entities. After investigating the reason behind this, we found that the Wikipedia pages corresponding to these entities do not exist any more.

\xhdr{Candidate generation} The starting point of most existing techniques in the entity linking literature is the existence of a large amount of high quality annotated data. This allows to reliably estimate the probability that a mention $m$ refers to an entity $e$, formally denoted as $P (e | m)$ and commonly referred to as \textit{prior} probability. Thus, a common choice for most of the works \cite{bhagavatula2015tabel,yamada,ganea,gupta2017entity,Kolitsas2018EndtoEndNE} is to leverage the prior information collected by \citeauthor{spitkovsky2012cross} \cite{spitkovsky2012cross}, who crawled the Web for associating strings of text with entities in the underlying English Wikipedia knowledge graph. This prior information constitutes the basis of the candidate generation system used by most entity linking techniques. It has been shown multiple times that prior information itself represents the bulk of the performance of the entity linking methods \cite{ratinov2011local,bhagavatula2015tabel,xel3,xel4}, and hence, some authors \cite{bhagavatula2015tabel,yamada} have spent much effort in crawling additional Web resources to complement \citeauthor{spitkovsky2012cross}'s corpus.

\begin{figure}[t]
    \centering
    \includegraphics[width=0.6\linewidth]{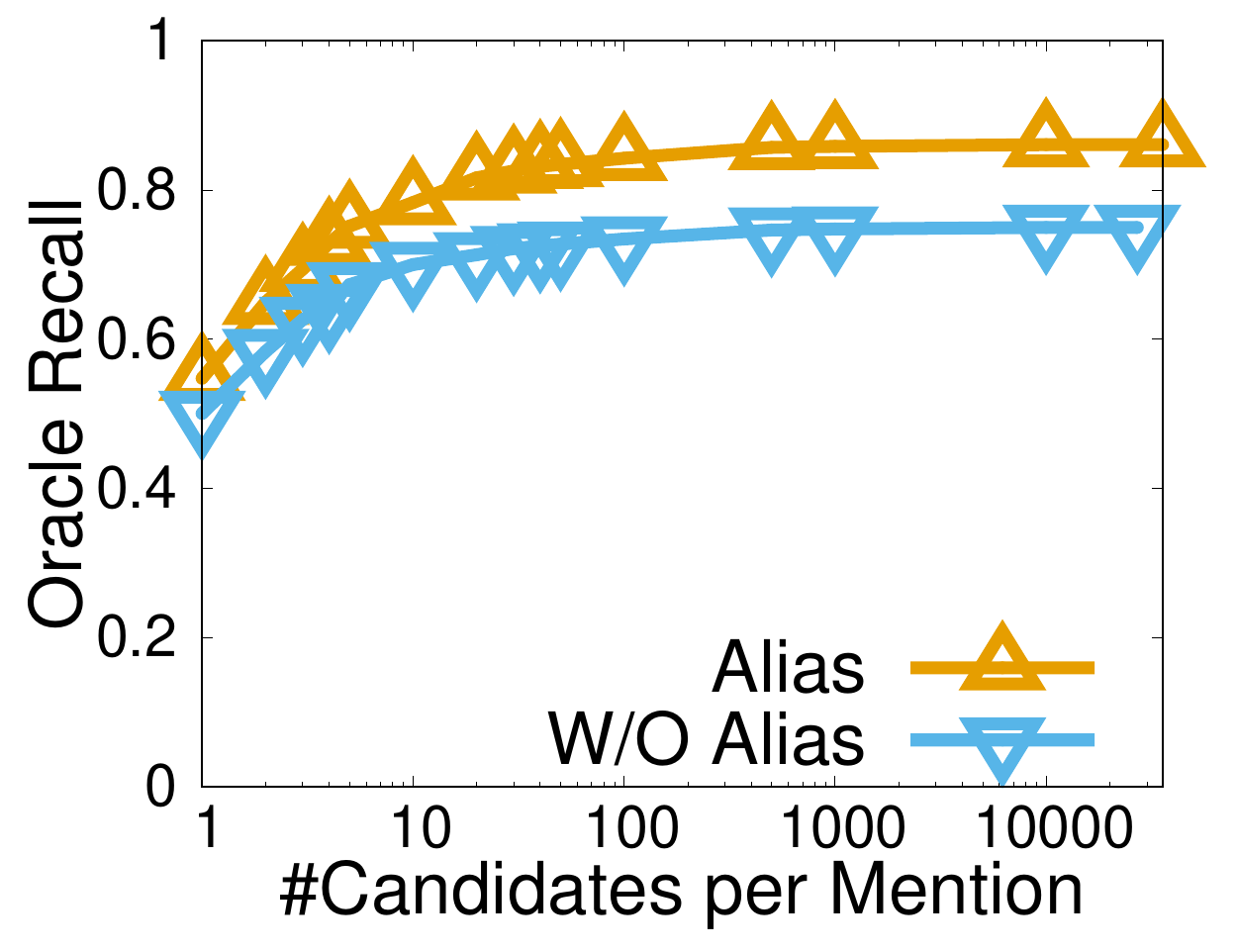}
    \figcaption{Oracle recall as a function of $T$ on the CoNLL-Val dataset (x-axis in log scale).}
    \label{fig:recall_conll_dev}
    \vspace*{1mm}
\end{figure}

However, in the absence of annotated data, such prior information is not available, and there is a requirement of mechanisms to generate a set of plausible candidate entities for a given mention string. To this end, and because we do not want to presume information more complex than what is readily available for arbitrary knowledge graphs, we employ a simple and efficient approach to generate candidates. For each entity $e$ in the knowledge graph $\mathcal{G}$, we have access to its name and a list of aliases used to commonly refer to $e$. Note that the mention strings, entity names, and aliases are tokenized into words. Given a mention $m$, the entities that contain all the tokens from $m$ are considered candidate entities. For example, \textsc{Michael Jordan (basketball player)} and \textsc{Michael Jordan (computer scientist)} are candidates for the mention \textsc{Michael Jordan}, while \textsc{Michael Jackson} is not. Although simple, the recall of this candidate generation approach is quite high and practical (as shown in Fig.~\ref{fig:recall_conll_dev}) for the data sets considered in this work. The methodology to perform disambiguation is orthogonal to the candidate generator, and while the latter could be improved \cite{sil2012linking,wang2015language}, which will only improve the performance of any technique, it is not in the scope of this work.
 
As the candidate generation system $\mathcal{C}_T$ only accounts for at most $T$ most likely entities of a mention, a sorting criterion is required. To this end, candidate entities are sorted based on the degree of their corresponding vertices in the undirected version of the knowledge graph. The degree of an entity roughly captures its popularity. 

A na\"{\i}ve implementation of the aforementioned string matching based candidate generator is impractical considering the sheer size of real-world knowledge graphs. To this end, we employ an \emph{inverted index} for scaling up the candidate generator. Specifically, for each token we maintain a set of entities containing that token in their name or in one of their aliases. Thus, for a mention string, we obtain the sets of entities corresponding to each of its constituent tokens, and the final set of candidates is computed by finding an intersection of the previously obtained sets. This ensures scalability and practicality of the candidate generation module.

We follow the aforementioned approach and use the `name' and `alias' information (available for 43.7M and 6.6M entities, respectively), described by the \texttt{<http://schema.org/name>} and \texttt{<http://www.w3.org/2004/02/skos/\linebreak core\#altLabel>} Wikidata relationships respectively, to construct an inverted index of tokens in the mentions to their corresponding Wikidata QIDs. This index is then used to generate candidate entities for each mention, where the entities are sorted based on their degree in Wikidata.

\xhdr{Fixing the maximum number of candidates per mention $T$} We analyze the effect of the parameter $T$ on the oracle recall for the CoNLL-Val dataset. As portrayed in Fig.~\ref{fig:recall_conll_dev}, the simple string matching based candidate generator achieves an overall oracle recall of $76\%$, which is improved to $87\%$ by incorporating the use of alias information. Furthermore, it is evident that increasing $T$ provides diminishing marginal gains in the oracle recall: retaining at most $20$ candidates per mention already results in an oracle recall of $83\%$, which is about $95\%$ of the overall oracle recall. With this observation and similar to the existing techniques in the literature \cite{ganea,titov_unsup}, we fix $T$ to be $20$ for all the datasets. While the candidate generator could be improved by employing smarter tokenization rules, fuzzy string matching, and using word embeddings to capture semantics, we believe the obtained oracle recall of $87\%$ is already quite high, as even with the use of prior information in the presence of annotated data the oracle recall for the aforementioned datasets are in the range of $90$ to $95\%$ \cite{ganea}.

\section{Experimental Setup}
\label{app:setup}
All the experiments were done using code written in Python on an Intel(R) Xeon(R) E5-2680 24-core machine with 2.50GHz CPU, and 256 GB RAM running Linux Ubuntu 20.04. For \ethemes and \wethemes, we use the open source implementation of the singular value decomposition and eigendecomposition, respectively, available in the NumPy linear algebra library \cite{numpy}. We adapt the Python implementation\footnote{\url{https://github.com/lephong/dl4el}} of \titov \cite{titov_unsup} made available by the authors themselves to work with Wikidata as the referent KB instead of Freebase. For details, please see Appendix~\ref{app:titov_implementation}.

\subsection{\textbf{Code and Datasets}}
\label{app:code_data}
The code, datasets, and all the other resources such as word and entity embeddings, required to reproduce the results reported in this paper are available at~\url{https://github.com/epfl-dlab/eigenthemes}.

\subsection{Note on the Implementation of \titov}
\label{app:titov_implementation}
\citeauthor{titov_unsup} \cite{titov_unsup} use Freebase as the referent KB. All information pertaining to candidate entities, such as entity types, name, aliases, description, popularity etc., was extracted from Freebase. To make the implementation of \titov compliant with our setup, while at the same time ensuring the use of their original resources as much as possible, we use the `Freebase ID', described by the property \texttt{<https://www.wikidata.org/wiki/\linebreak Property:P646>}, available in the Wikidata dump to map Freebase entities to Wikidata entities. This enabled us to identify a mapping from Freebase to Wikidata for approximately $1.3$M entities. After obtaining the mapping, it was straightforward to adapt the implementation of \titov to work on our preprocessed datasets (Sec.~\ref{subsec:setup}) using their original resources. We provide modified source files for \titov's implementation in our GitHub repository~\url{https://github.com/epfl-dlab/eigenthemes}, which can be simply used to replace the corresponding files at~\url{https://github.com/lephong/dl4el} to obtain the results for \titov reported in this paper. Detailed instructions are provided in the README of our GitHub repository.

\begin{figure*}[t]
\moveup
\moveups
\centering
    \subfloat[Micro precision@1]
    {
        \scalebox{0.23}{
            \includegraphics[width=\linewidth]{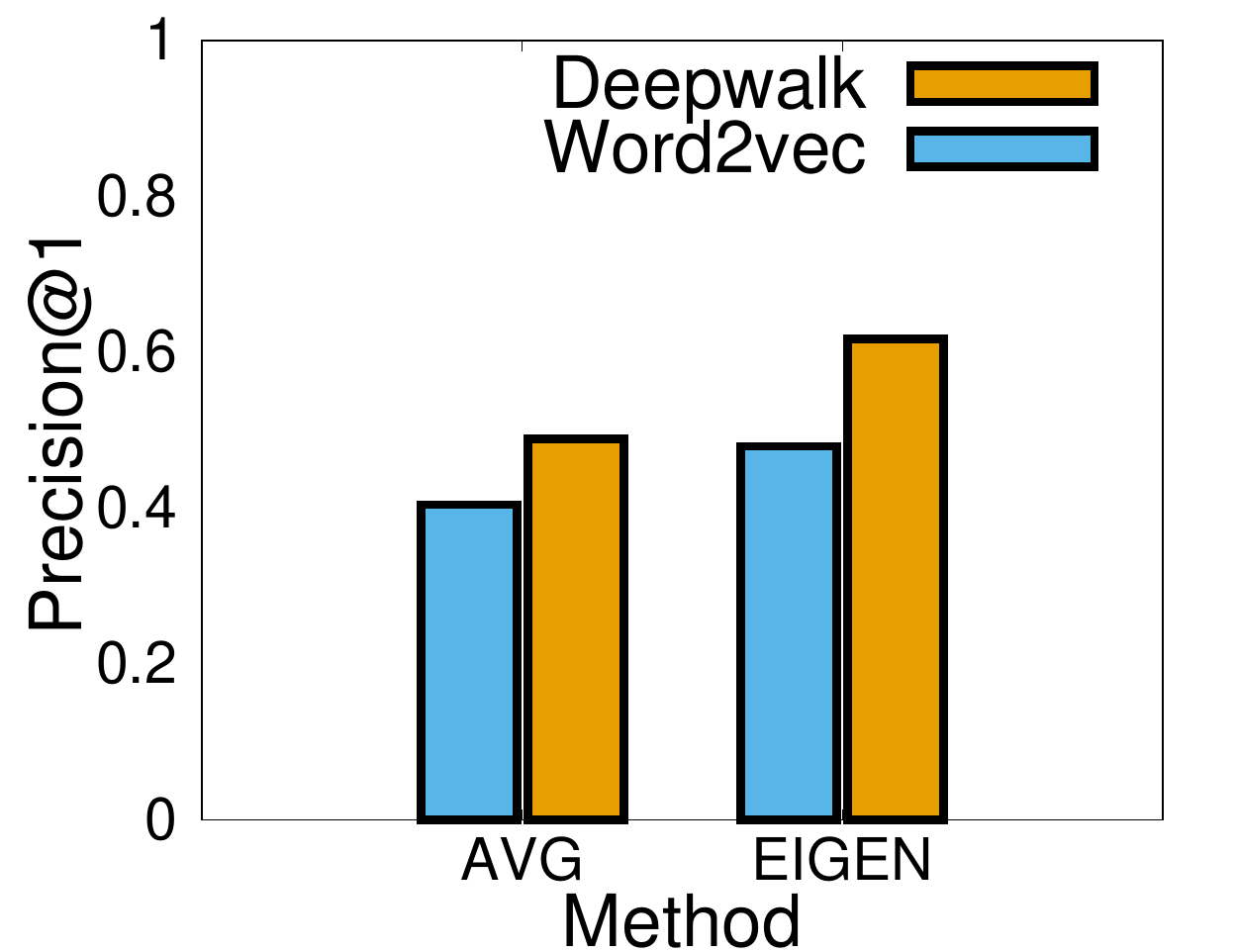}
        }
        \label{fig:embedding_choice_precision_dev}
    }
    \subfloat[Micro MRR]
    {
        \scalebox{0.23}{
            \includegraphics[width=\linewidth]{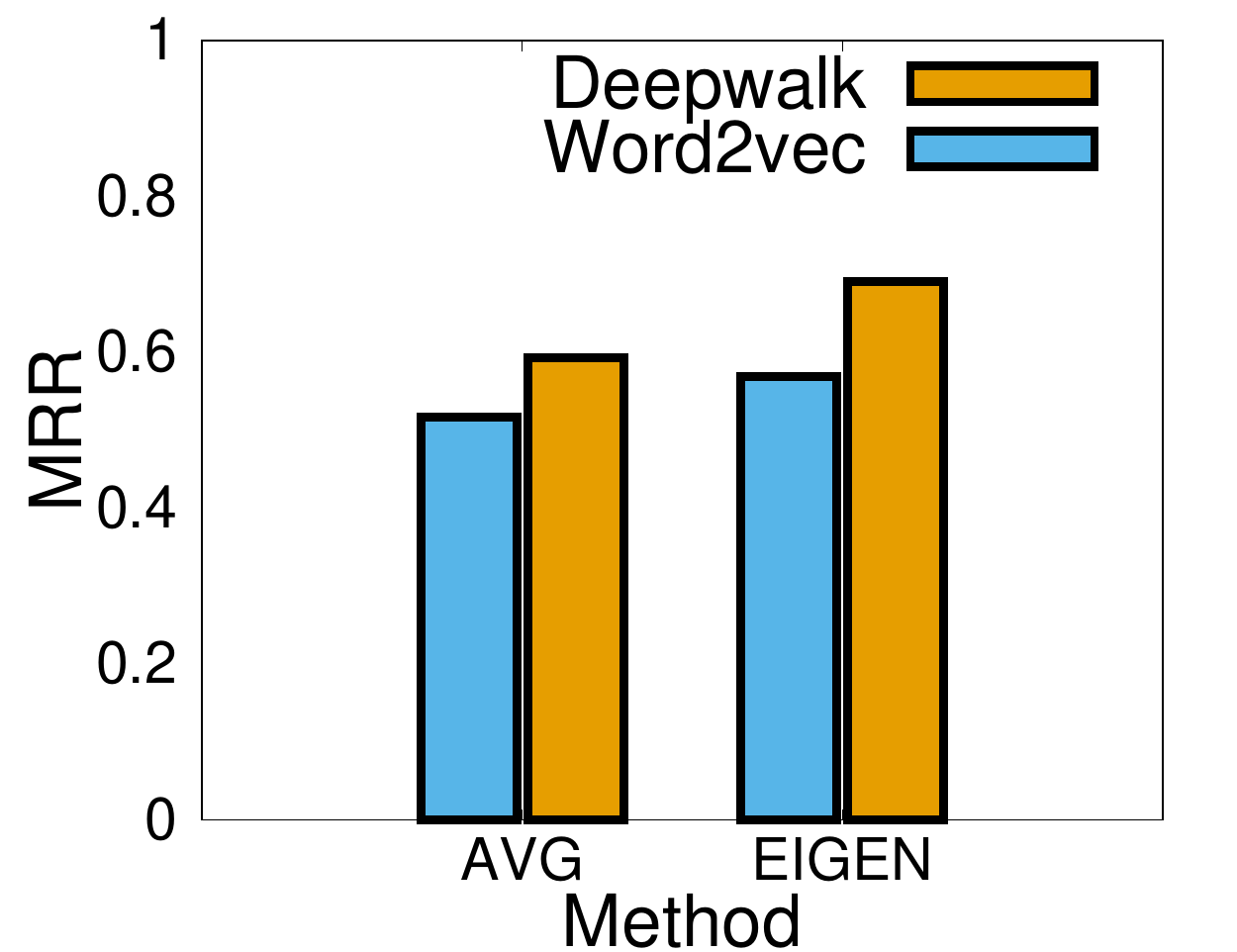}
        }
        \label{fig:embedding_choice_mrr_dev}
    }
    \subfloat[Micro precision@1]
    {
        \scalebox{0.23}{
            \includegraphics[width=\linewidth]{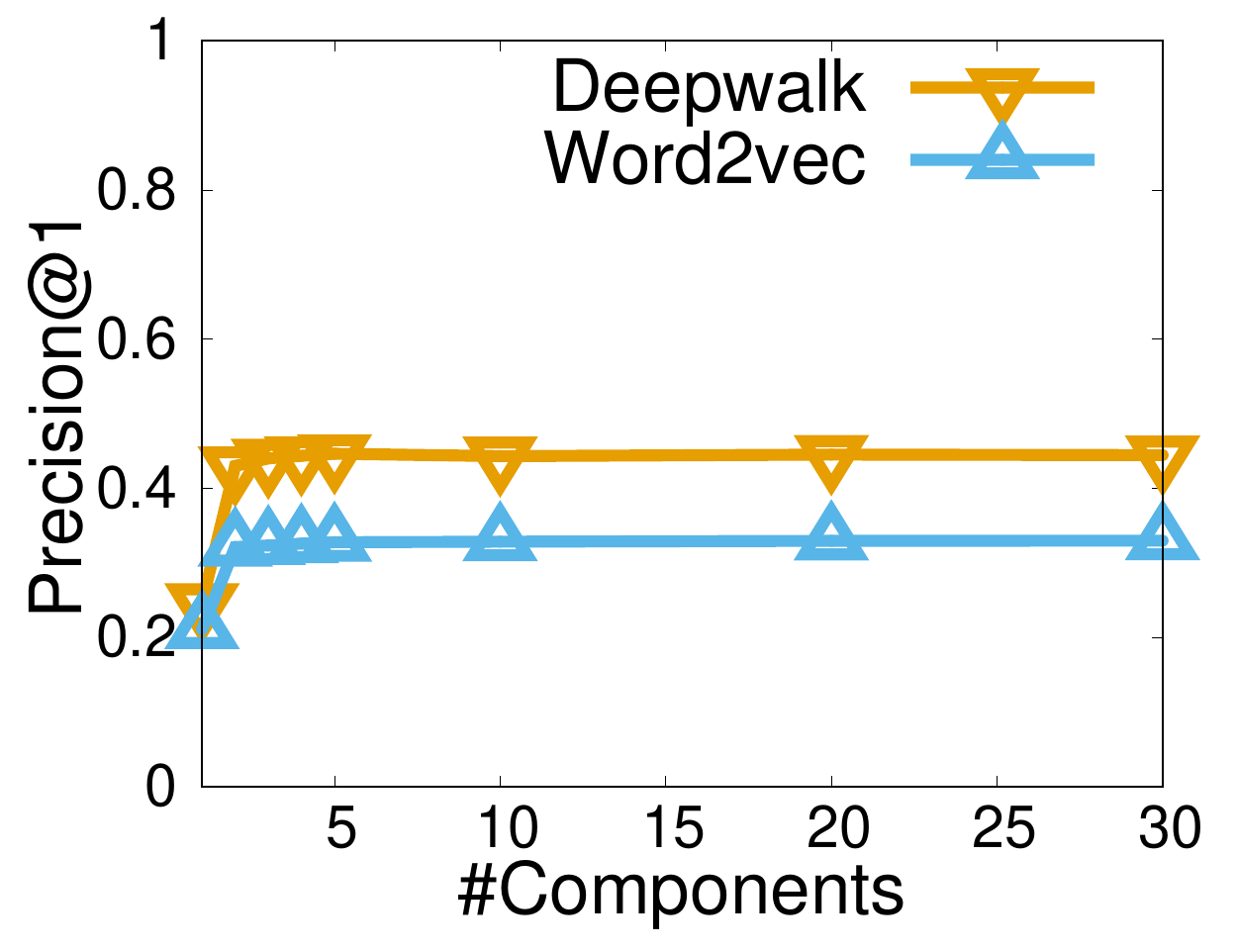}
        }
        \label{fig:comps_precision_dev}
    }
    \subfloat[Micro MRR]
    {
        \scalebox{0.23}{
            \includegraphics[width=\linewidth]{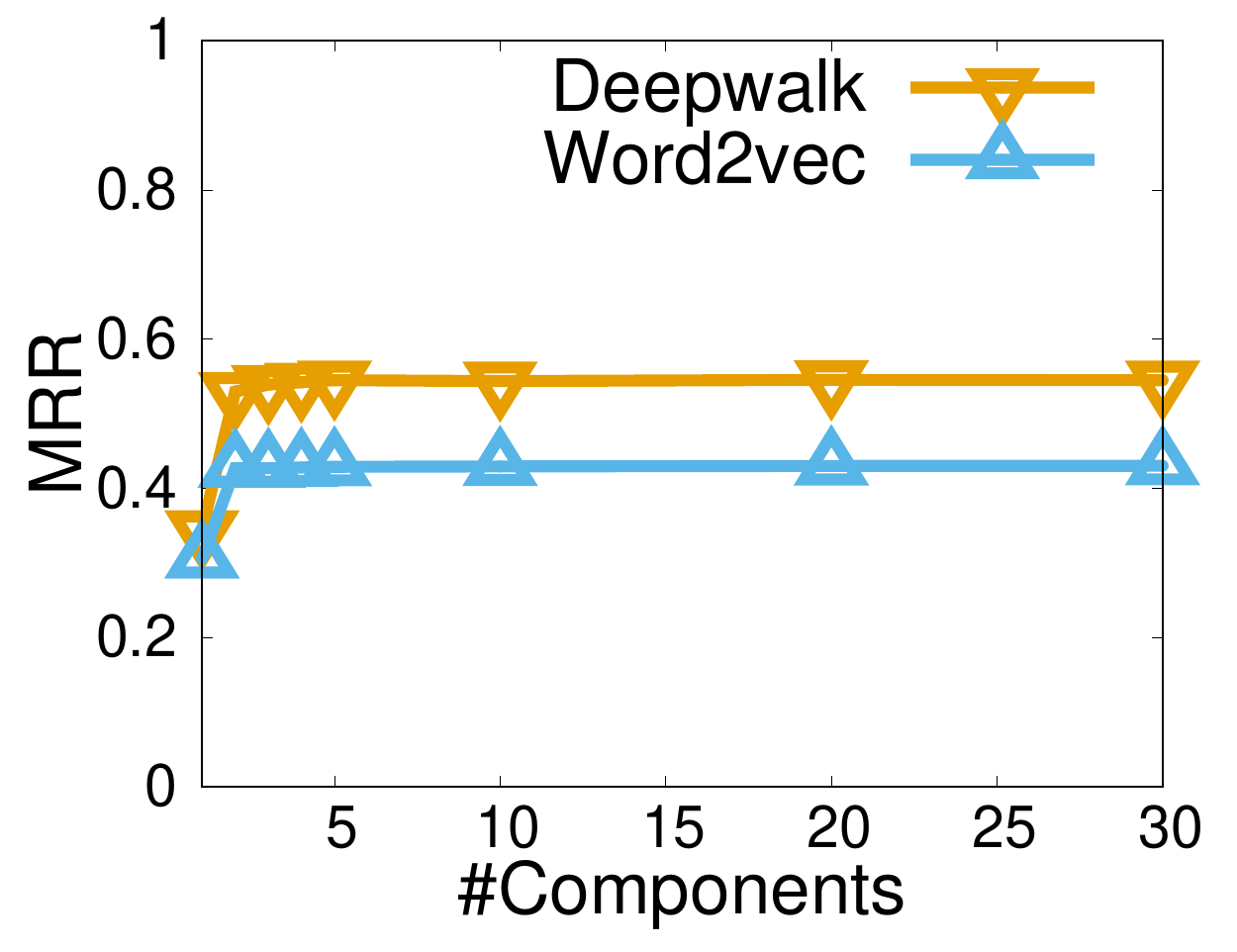}
        }
        \label{fig:comps_mrr_dev}
    }
    \figcaption{Analyzing the effect of (a)-(b) the method for obtaining entity embeddings, and (c)-(d) the number of components used to construct the low-rank subspace using \eshort, on EL quality for the CoNLL-Val dataset.}
\label{fig:param_tuning}
\vspace*{1mm}
\end{figure*}

\begin{figure*}[t]
\moveup
\centering
    \subfloat[Weights: Precision@1]
    {
        \scalebox{0.23}{
            \includegraphics[width=\linewidth]{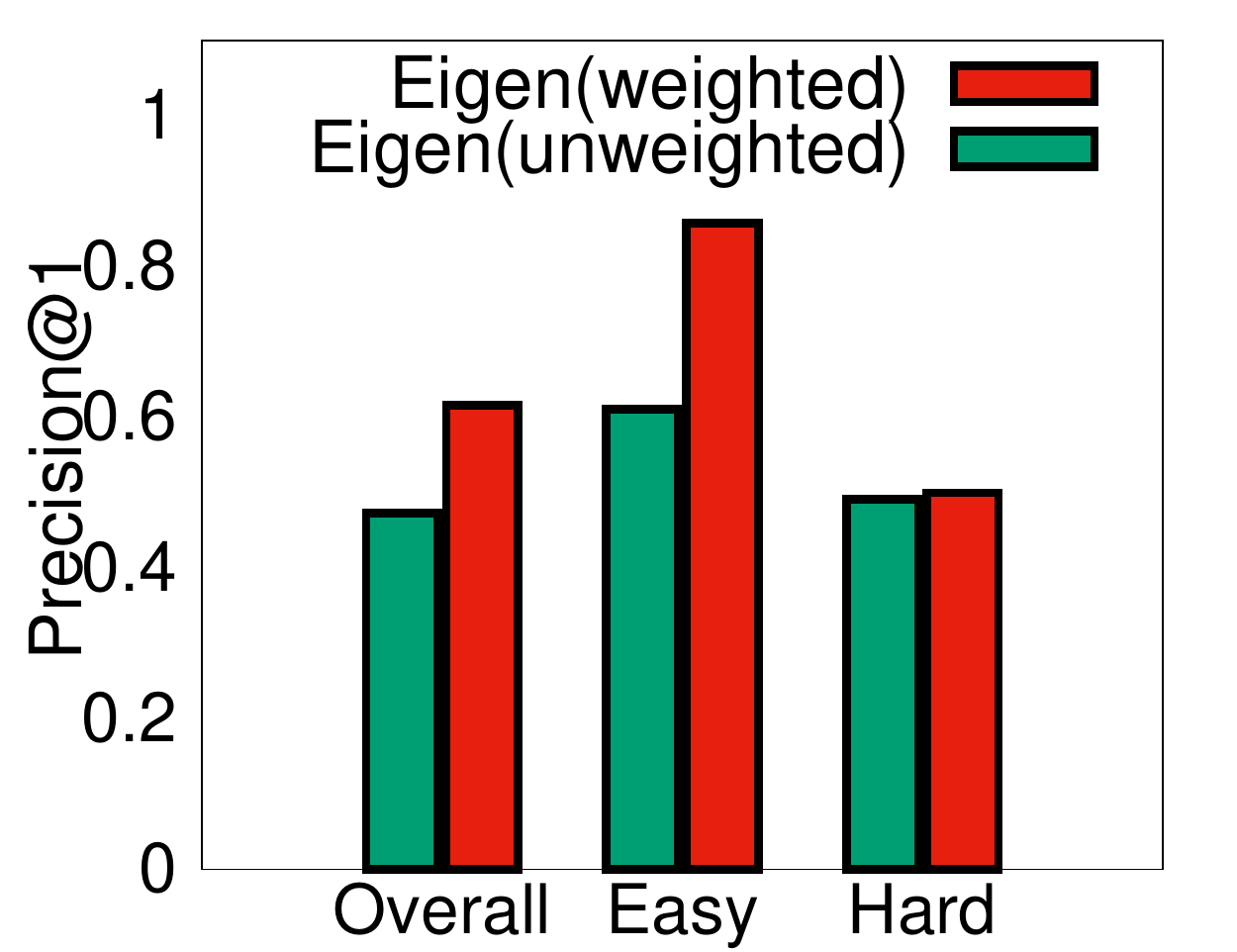}
        }
        \label{fig:weights_precision}
    }
    \subfloat[Weights: MRR]
    {
        \scalebox{0.23}{
            \includegraphics[width=\linewidth]{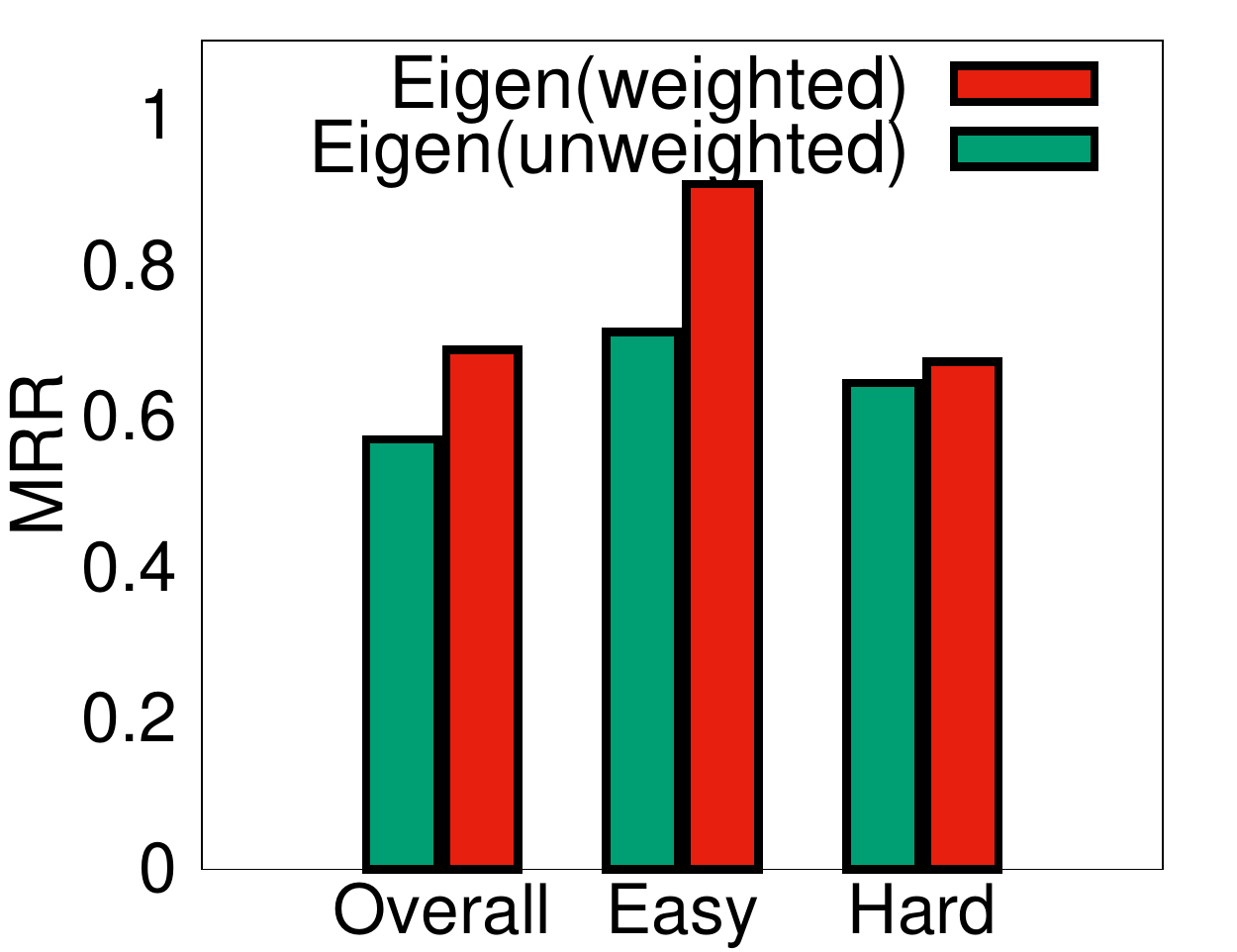}
        }
        \label{fig:weights_mrr}
    }
    \subfloat[WScheme: Precision@1]
    {
        \scalebox{0.23}{
            \includegraphics[width=\linewidth]{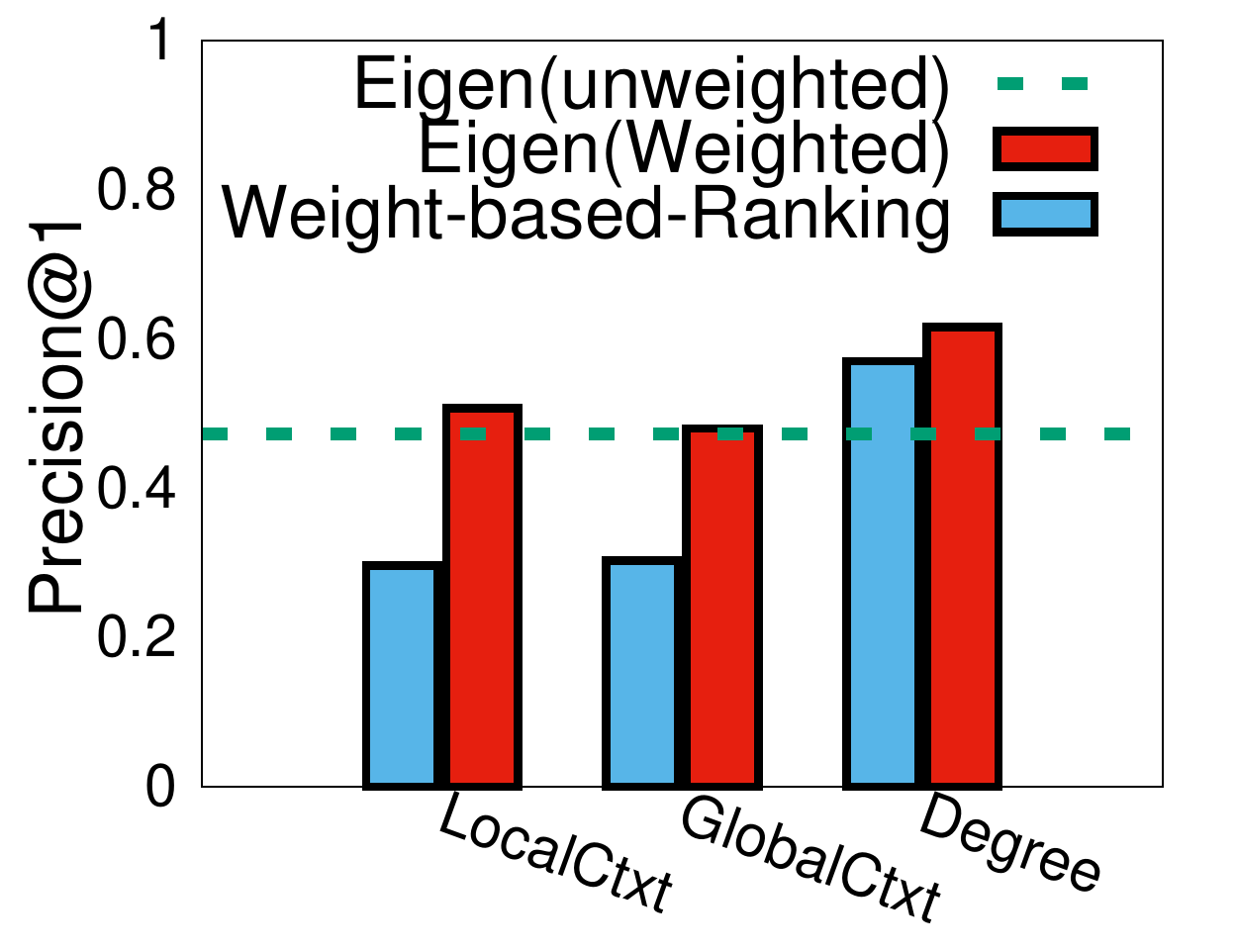}
        }
        \label{fig:weighting_scheme_precision}
    }
    \subfloat[WScheme: MRR]
    {
        \scalebox{0.23}{
            \includegraphics[width=\linewidth]{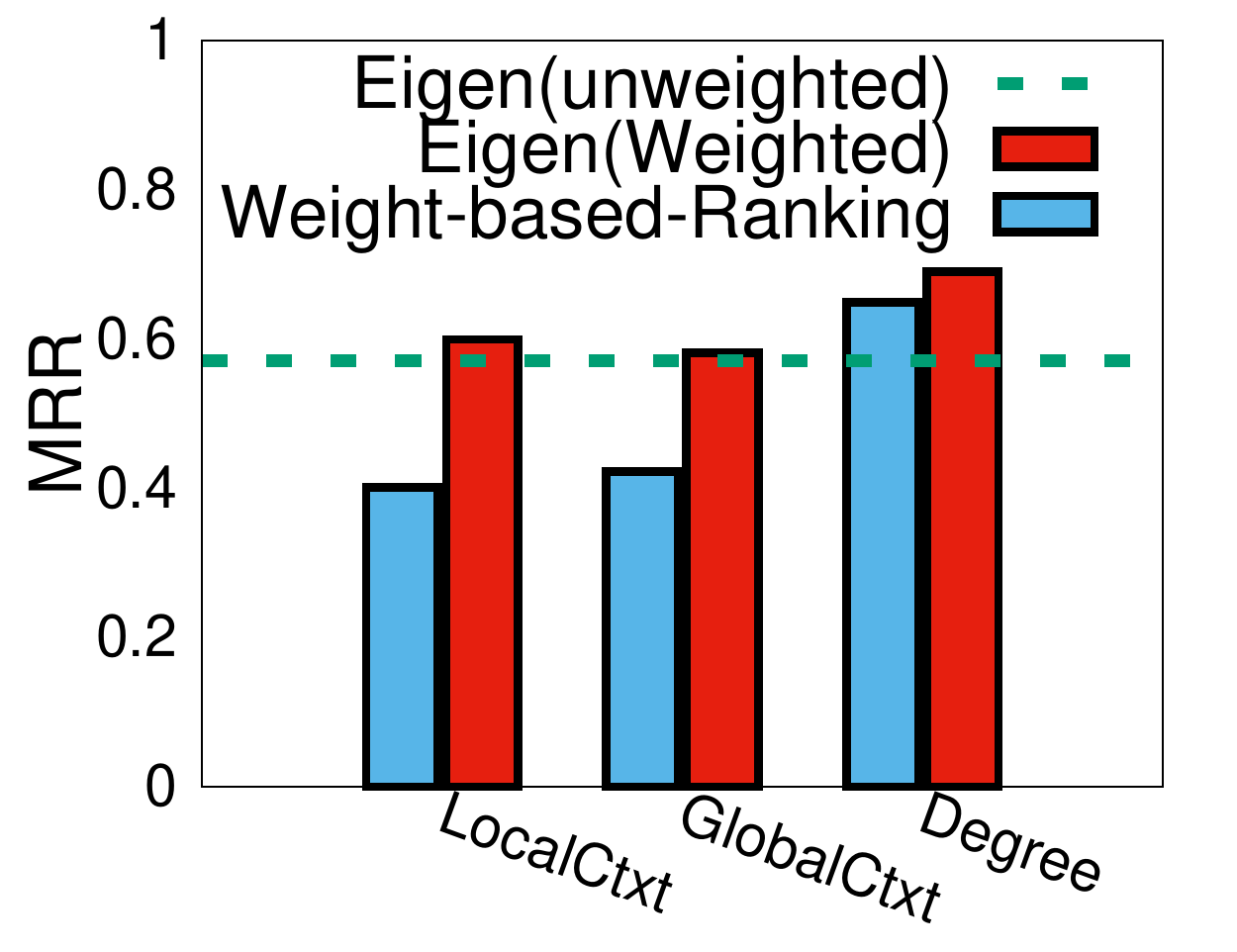}
        }
        \label{fig:weighting_scheme_mrr}
    }
    \figcaption{Comparing entity linking quality of unweighted and weighted \eshort using (a) precision@1, and (b) MRR on the CoNLL-Test dataset. Analyzing the effect of the weighting scheme for constructing the weighted subspace on the entity linking performance of weighted \eshort using (c) precision@1, and (d) MRR for the CoNLL-Test dataset.}
\label{fig:analysis_weights_app}
\end{figure*}

\subsection{Entity Embeddings}
\label{app:entity_embed}
The entity embeddings can be computed using the following two approaches:

\noindent $\bullet$ \textbf{Graph structure:} This approach presents a natural way of learning entity embeddings by building models that preserve the neighborhood of entities in an underlying graph. To this end, we construct a subgraph of the Wikidata knowledge graph by retaining only the edges existent between Wikidata entities that have a mapping in the English Wikipedia. The entity embeddings are then learned by training \dw \cite{deepwalk} on the resultant graph of 3.7M entities and 20.2M edges. 

We also explored recent state of the art techniques NetSMF \cite{netsmf} and LouvainNE \cite{louvain} for learning entity embeddings using the Wikidata graph. While NetSMF \cite{netsmf} crashed on our machine (with 256 GB RAM) owing to going out of memory, LouvainNE \cite{louvain} was much worse than \dw on the entity relatedness task. Specifically, while \dw obtained an MRR of 0.62, LouvainNE obtained that of 0.44. Note that the authors of LouvainNE did not conduct any experiments on the entity relatedness task, and while it was shown to be better than \dw on the node classification task, it cannot be used, since the ability to better capture entity relatedness is a desired property for any embedding technique to be useful in our setting.

\noindent $\bullet$ \textbf{Textual descriptions:} KGs usually offer a short textual description for each entity and Wikidata is no exception. Entity descriptions have been used to learn entity embeddings in the literature \cite{Yamada2018RepresentationLO}. In the same vein, we learn entity embeddings by extracting entity descriptions provided by the \texttt{<http://schema.org/description>} Wikidata relationship, and computing the average of \wv embeddings of the description words.

Moving ahead, we study the utility of embeddings obtained by the aforementioned approaches on our downstream entity linking task. A recent study \cite{Almasian2019WordEF} indicates the ability of node embedding based methods to better capture relatedness when compared to word embedding based methods, which are instead better in capturing similarity. Since the objective of subspace learning is to capture topical relatedness across the gold entities in a document $D$, the subspace $\mathcal{S}_D$ learned over the candidate embedding matrix $\mathbf{E}_D$ constructed using graph-structure based embeddings should perform better than those obtained using textual descriptions.

\xhdr{Results} We also empirically validate the aforementioned intuition. Figs.~\ref{fig:embedding_choice_precision_dev} and~\ref{fig:embedding_choice_mrr_dev} show that there exists a stark difference in the entity linking quality (of around $8$ to $12$ percentage points in both precision@1 and MRR) for both \avg and \eshort when using entity embeddings obtained by \dw than those obtained by average of \wv embeddings of the entity description words. This observation provides substantial evidence in favor of using \dw over \wv embeddings for learning the subspace representation. Therefore, we fix the method for obtaining entity embeddings to the graph-structure based approach using \dw.

\subsection{Weighting Scheme}
\label{app:weighting}
The weighting scheme  $\mathcal{W}$ establishes a descending sorting that relates to the (presumable) probability of each of the candidate entities being linked to the mention under consideration. 

Let $\mathcal{W}(e)$ denote the weight for the candidate entity $e$, which is computed as follows:
\begin{equation}
    \mathcal{W}(e) = \text{rank}(e)^{-\delta}
    \label{eq:weight}
    \moveups
\end{equation}
where rank($e$) corresponds to the position of the entity $e$ in the ranking computed. This is closely related to the reciprocal rank, which is a widely used in the information retrieval community. The parameter $\delta$ ($>0$) controls the importance of the rank position. For large values of $\delta$, the weights will decay very quickly with respect to the rank position, whereas for small values the weights will become more uniform. 

A ranking is established for the candidate entities of each mention. Thus, for a document with $5$ mentions we have 5 rankings. We explore two different ranking mechanisms.

\noindent\textbf{Ranking based on entity degree.}
This ranking relies on graph information. It simply takes the order provided---based on entity degree in the knowledge graph---by the candidate generation system, and uses that ranking to compute the weights.

\noindent\textbf{Ranking based on textual coherence.}
This ranking leverages text signals. For each entity of the knowledge graph, a small description is available. Examples of these descriptions are ``British author and humorist'' or ``Republic in Southwestern Europe'' for the entities \textit{Douglas Adams} and \textit{Portugal}, respectively. Similar to \cite{Yamada2018RepresentationLO}, we use pre-trained word embeddings to represent entities as the average of the embeddings of the words in their description. We also use these word embeddings to compute a context representation as the average of the embeddings of the words that surround the mention under consideration. We then compute the cosine similarity between the description embedding of each of the candidate entities and the context embedding. The ranking is established based on these similarity scores. For the context embedding we consider a \textbf{local} context, which only takes into account words within a window size from the mention, and a \textbf{global} context, which takes into account all the words in the document.

\xhdrNoPeriod{Analysis: Do weights enrich the quality of subspaces learned by \eshort?}
We empirically assess the utility of the extension (Sec.~\ref{sec:weighted-eigenthemes}) for incorporating weights in the subspace learning step of \eshort. It is evident from Figs.~\ref{fig:weights_precision} and~\ref{fig:weights_mrr} that \eshort (with weights) learns improved subspace representations and consequently, obtains better entity linking performance than \eshort (without weights). Furthermore, while \eshort (weighted) is only marginally better than \eshort (unweighted) for `hard' mentions, it is substantially better (about $25$ percentage points) for `easy' mentions. This is easily explained as the signal derived from the entity-degree-based weighting scheme is biased towards easy mentions by construction (Sec.~\ref{subsec:data}). Thus, \eshort with weighting is expected to obtain an improvement in quality for the easy mentions. In fact, the simultaneous (though marginal) improvement obtained for hard mentions indicates that \eshort (weighted) successfully congregates the best of both the worlds. This also reinforces the robustness of the \ethemes framework, as the learned weighted subspace did not get biased towards improving the performance for easy mentions alone. Similar to \eshort, weights also improve \avg, \ie, \avg (weighted) is considerably better (about $10$ points) than \avg (unweighted). 

\xhdrNoPeriod{Analysis: Which weighting scheme is the most efficacious?}
We now empirically assess the strength of different weighting schemes towards improving the EL performance of other techniques, and use \eshort as a representative technique for the analysis. Results for other techniques portray similar trends and are omitted for the sake of brevity. Figs.~\ref{fig:weighting_scheme_precision} and~\ref{fig:weighting_scheme_mrr} present a comparison of the quality of three signals: (1) \lctxt, and (2) \gctxt (text-based), and (3) entity degree (graph-based); and their impact on the quality of \eshort (weighted). It is evident from the bar corresponding to `Weight-based Ranking' in both Figs.~\ref{fig:weighting_scheme_precision} and~\ref{fig:weighting_scheme_mrr}, \ie, the ranking of candidates obtained solely using the weighting schemes, that the text-based signals are much weaker than the graph-based signal. Hence, not surprisingly, the quality of \eshort (weighted) using text signals is only marginally better than the unweighted \eshort. On the other hand, the graph-based signal results in a substantial improvement of around $14$ percentage points for \eshort (weighted) over \eshort (unweighted). Thus, we choose the ranking induced by entity degree as the preferred weighting scheme.

\begin{figure}[t]
\moveup
\centering
    \subfloat[$T$ and $\delta$: Precision@1]
    {
        \scalebox{0.46}{
            \includegraphics[width=\linewidth]{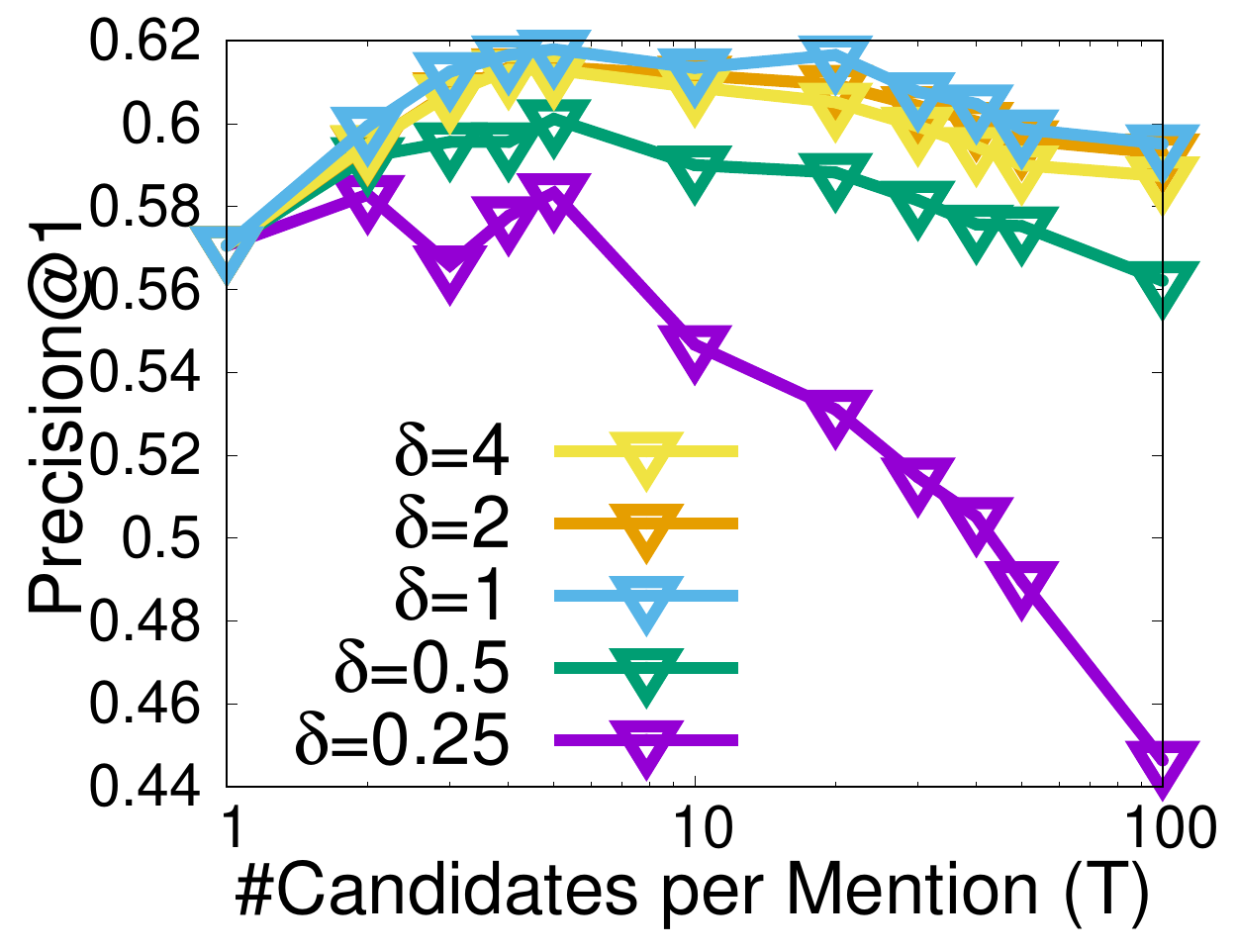}
        }
        \label{fig:ncands_wpca_precision_test}
    }
    \subfloat[$T$ and $\delta$: MRR]
    {
        \scalebox{0.46}{
            \includegraphics[width=\linewidth]{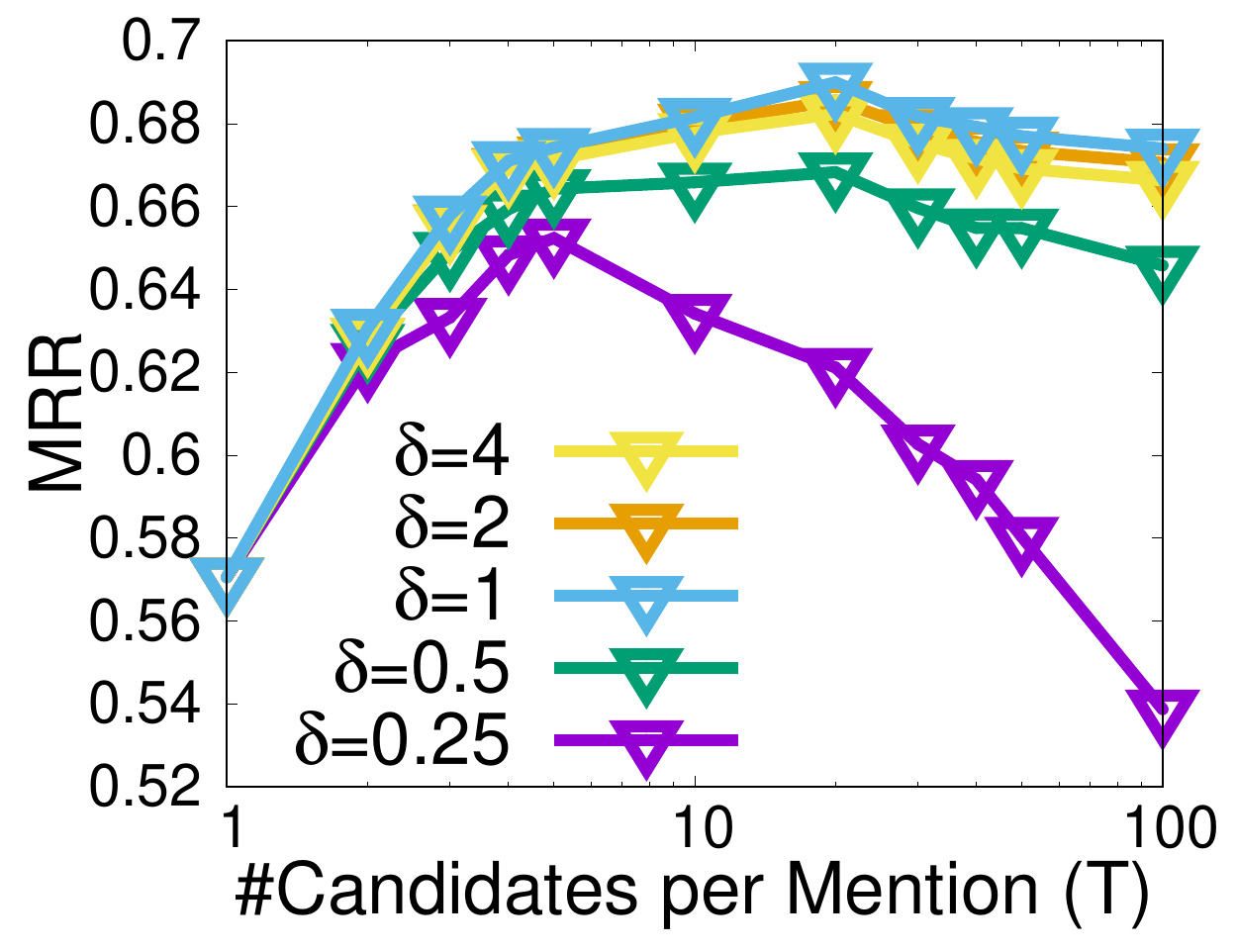}
        }
        \label{fig:ncands_wpca_mrr_test}
    }
    \figcaption{Analyzing the effect of the maximum number of candidates per mention ($T$) on the entity linking performance of \eshort using (a) precision@1, and (b) MRR for the CoNLL-Test dataset.}
\label{fig:analysis_ncands_app}
\end{figure}

\xhdr{Analysis: Effect of $\delta$} Moving ahead, we also analyze the effect of the parameter $\delta$, which reflects the intensity of the weights (Eq.~\ref{eq:weight}) in the subspace learning step of \eshort. For any fixed value of $T$, Figs.~\ref{fig:ncands_wpca_precision_test}--~\ref{fig:ncands_wpca_mrr_test} portray that increasing $\delta$, and therefore the intensity of weights, results in an improvement at first (until $\delta=1$), beyond which the performance starts to slowly decline. An in-depth analysis revealed that increasing $\delta$ always results in an improvement for easy mentions, however, the performance on hard mentions improves until $\delta=1$ while starts to decline beyond that. We will see later in Appendix~\ref{app:disparity_easy_hard}, that \degree (by construction) possesses a bias towards easy mentions, 
and this is exactly what is at play here. To summarize, beyond $\delta=1$ \eshort gets biased towards easy mentions and thus, the performance on hard mentions starts to decline, which is also reflected marginally in the overall performance. More elaborate weighting schemes might facilitate an even larger improvement in the performance of \eshort, however, we leave the design of such schemes for future work.

\xhdr{Analysis: Effect of the number of candidates} While in the experiments, $T$ was fixed to $20$ based on the analysis in Appendix~\ref{app:preprocess}, we now analyze the effect of variation in this parameter on the performance of \eshort. Figs.~\ref{fig:ncands_wpca_precision_test}--~\ref{fig:ncands_wpca_mrr_test} show that the performance of \eshort improves at first (up to $5$ candidates per mention), beyond which the performance starts to slowly decline (barring $\delta=0.25$ for which the decline is steep). This is because increasing the number of candidates beyond a point makes the document embedding matrix noisy, thereby affecting the quality of the learned subspace. However, it can be observed that for larger values of $\delta\geq1$, \eshort gracefully handles the noise resulting from an increase in the number of candidates, and we only observe minor effects on the EL performance.

\subsection{Results CoNLL: `Easy' \vs `Hard'}
\label{app:disparity_easy_hard}
We found that the average number of candidates per mention for hard mentions is much higher than that for easy mentions, and this phenomenon has a stronger deteriorating effect on \titov than \eshort. Consequently, the improvement (around $30$ points) of \eshort over \titov is even more profound for the hard mentions. Note that \degree obtains a performance of $100\%$ for easy and $0\%$ for hard mentions, which is simply due to the way these sets were constructed (\cf Sec.~\ref{subsec:data}).

\section{Hyper-parameter Tuning}
\label{app:tuning}
The only hyperparameter for \eshort is the number $k$ of components used to construct the low-rank subspace representation of each document $D$. In this section, we analyze the effect of $k$ on the quality of the entity linking output of \eshort (unweighted) using the CoNLL-Val dataset. Note that the outcome of the analyses presented in this section is generalizable to \eshort (weighted) as it is a specialized instance of \eshort (unweighted). Naturally, we observe similar trends for \eshort (weighted) and the results are therefore omitted.

It is evident from both Figs.~\ref{fig:comps_precision_dev} and~\ref{fig:comps_mrr_dev} that increasing the number of components results in an improvement in the entity linking quality measured using precision@1 and MRR respectively. Furthermore, the performance improves monotonically and then plateaus around $10$ components, post which there is no considerable improvement. Therefore, we fix the number of components $k$ to $10$ for obtaining all the experimental results presented in this paper.

\begin{table}[t]
\vspace*{3mm}
\tabcaption{Supervised: precision@1 for CoNLL-Test. We run each model $5$ times and report the mean and standard deviation.}
\vspace*{1mm}
\label{tab:supervised}
\centering
\scalebox{0.88}{
\begin{tabular}{c||c|c|c}
\hline
\bf Context & \multirow{2}{*}{\bf F1+F2+F3} & \multirow{2}{*}{\bf F1+F2+F3+F4} & \multirow{2}{*}{\bf Ceiling} \\
\bf Type & & & \\
\hline
\textbf{Local} & 0.848 $\pm$ 0.002 & \bf{0.864 $\pm$ 0.0003} & 0.974 \\
\hline
\textbf{Global} & 0.852 $\pm$ 0.002 & \bf{0.862 $\pm$ 0.002} & 0.974 \\
\hline
\end{tabular}
}
\moveup
\moveup
\end{table}

\subsection{Extraneous Parameters}
\label{app:params}
We use the parameters prescribed in \cite{deepwalk} to train \dw, with the number and length of random walks per node set to $80$, and the dimensionality of entity embeddings set to $128$. Unless stated otherwise, the `local' textual context of each word in a document $D$ is computed as the average of the word embeddings of its $5$ surrounding words (to the left and right), while the `global' context is computed as the average of the embeddings of all the nouns \cite{yamada} in $D$. The Apache OpenNLP tagger \cite{tagger} was used to detect nouns.

\section{\ethemes for Supervised Settings}
\label{app:supervised}
Having established the superiority of \ethemes on the CoNLL-Test dataset for the unsupervised setting, we now assess their applicability in settings where annotated data is available. Note that the goal of this experiment is not to design state of the art supervised entity linking systems. Rather, the focus is to showcase the capability of \ethemes to improve them. 

We first describe the setup, which is slightly different from the unsupervised case. For candidate generation, we use the mention--entity association dictionaries made available publicly by \citeauthor{ganea} \cite{ganea}. Similar to the unsupervised case, $T$ was fixed to $20$. In addition to facilitating candidate generation, this dictionary allows us to infer the prior probability $P (e | m)$. The availability of annotated data also allows for the learning of aligned word and entity embeddings. We employ $300$-dimensional pre-trained, aligned entity and word embeddings \cite{ganea}. Inspired by existing state of the art methods for supervised entity linking \cite{ganea, yamada}, for each mention--candidate pair, we use the following features to train a supervised model: (F1) the prior probability; (F2) textual context score: obtained by computing the cosine similarity between the candidate entity embedding and the (local or global) context embedding (Sec.~\ref{app:weighting}); (F3) global coherence score: obtained by computing the cosine similarity between the candidate entity embedding and the global entity context \cite{yamada}; and (F4) the \eshort score (Sec.~\ref{sec:similarity-function}).

We employ random forests \cite{rf} as a point-wise learning-to-rank technique to appropriately combine the contribution of the aforementioned features. We rely on the publicly available implementation of random forests in scikit-learn \cite{sklearn}. The model is trained using the CoNLL-Train set, while the CoNLL-Test set is used to evaluate the entity linking quality. The results are presented in Table~\ref{tab:supervised}. Adding the \eshort score as a feature into the supervised model results in an improvement of $1$ to $2$ percentage points. This result portrays the ability of our method to improve existing supervised entity linking systems. Furthermore, it validates the importance of an appropriate collective disambiguation method, even in the presence of other local scores such as prior information and contextual cues. Lastly, it is worth highlighting that the bulk of the improvement (more than $5$ percentage points) is obtained for the hard mentions, which is both required and important as existing features already facilitate obtaining the performance of close to $100\%$ on easy mentions.

\end{document}